\documentclass[sn-vancouver,Numbered]{sn-jnl}

\usepackage{graphicx}%
\usepackage{multirow}%
\usepackage{amsmath,amssymb,amsfonts}%
\usepackage{amsthm}%
\usepackage{mathrsfs}%
\usepackage[title]{appendix}%
\usepackage{xcolor}%
\usepackage{textcomp}%
\usepackage{manyfoot}%
\usepackage{booktabs}%
\usepackage{algorithm}%
\usepackage{algorithmicx}%
\usepackage{algpseudocode}%
\usepackage{listings}%
\usepackage{natbib}
\usepackage{float}
\usepackage[normalem]{ulem} 
\usepackage{graphicx}
\usepackage{array}
\usepackage{tabularx} 
\usepackage{soul}

\theoremstyle{thmstyleone}%
\theoremstyle{thmstyletwo}%

\theoremstyle{thmstylethree}%

\begin{document}

\title[Article Title]{Linguistic Entity Masking to Improve Cross-Lingual Representation of Multilingual Language Models for Low-Resource Languages}

\author*[1]{\fnm{Aloka} \sur{Fernando}}\email{alokaf@cse.mrt.ac.lk}

\author[2]{\fnm{Surangika} \sur{Ranathunga}}\email{s.ranathunga@massey.ac.nz}

\affil[1]{\orgdiv{Dept. of Computer Science \& Engineering}, \orgname{University of Moratuwa}, \orgaddress{\postcode{10400}
\country{Sri Lanka}}}

\affil[2]{\orgname{School of Mathematical and Computational Sciences, Massey University}, \orgaddress{\street{Palmerston North},  \postcode{4443}, \country{New Zealand}}}


\abstract{Multilingual Pre-trained Language models (multiPLMs), trained on the Masked Language Modelling (MLM) objective are commonly being used for cross-lingual tasks such as bitext mining. However, the performance of these models is still suboptimal for low-resource languages (LRLs). To improve the language representation of a given multiPLM, it is possible to further pre-train it. This is known as continual pre-training. Previous research has shown that continual pre-training with MLM and subsequently with Translation Language Modelling (TLM) improves the cross-lingual representation of multiPLMs. However, during masking, both MLM and TLM give equal weight to all tokens in the input sequence, irrespective of the linguistic properties of the tokens. In this paper, we introduce a novel masking strategy, \textit{Linguistic Entity Masking} (LEM) to be used in the continual pre-training step to further improve the cross-lingual representations of existing multiPLMs. In contrast to MLM and TLM, LEM limits masking to the linguistic entity types nouns, verbs and named entities, which hold a higher prominence in a sentence. Secondly, we limit masking to a single token within the linguistic entity span thus keeping more context, whereas, in MLM and TLM, tokens are masked randomly.  We evaluate the effectiveness of LEM using three downstream tasks, namely bitext mining, parallel data curation and code-mixed sentiment analysis using three low-resource language pairs English-Sinhala, English-Tamil, and Sinhala-Tamil. Experiment results show that continually pre-training a multiPLM with LEM outperforms a multiPLM continually pre-trained with MLM+TLM for all three tasks.}




\maketitle

\section{Introduction}\label{sec_intro}

Encoder-based Multilingual Pre-trained Language Models (multiPLMs) such as mBERT~\citep{devlin2019bert} and XLM-R~\citep{conneau2020xlmr} produce State-of-the-art results for many Natural Language Processing (NLP) tasks in the context of low-resource languages (LRLs)~\cite{acs2021evaluating, dhananjaya2022bertifying}. One success factor of these multiPLMs is the training objective utilized during the pre-training stage. mBERT and XLM-R were pre-trained using the Masked Language Modeling (MLM) objective. However, the performance of these models for cross-lingual tasks such as bitext mining had been suboptimal~\citep{hu2020xtreme}, due to the lack of an explicit cross-lingual pre-training objective~\cite {hu2021amber}. Translation Language Modeling (TLM) objective~\citep{conneau2019cross} was introduced to improve the cross-lingual capability of the existing multiPLMs. In contrast to MLM that uses only monolingual data, TLM uses parallel data across multiple languages in a \textit{continual pre-training} step to further improve the cross-lingual representations.~\citet{conneau2019cross} had proven that TLM improved the performance of cross-lingual classification and unsupervised Neural Machine Translation (NMT).

\begin{figure}[h]
    \caption{Self-attention weights among the words for an English and its corresponding Sinhala sentence. The darker the colour is, the stronger the relationship (ie. self-attention weight) between the two words. }\label{fig:attention_matrix}
    \centering
    \begin{minipage}[t]{0.5\textwidth}         \centering
        \includegraphics[width=\textwidth]{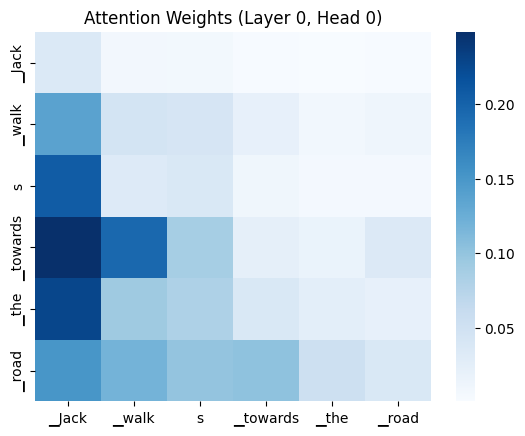}
    \end{minipage}%
    \begin{minipage}[t]{0.5\textwidth}
        \centering
        \includegraphics[width=\textwidth]{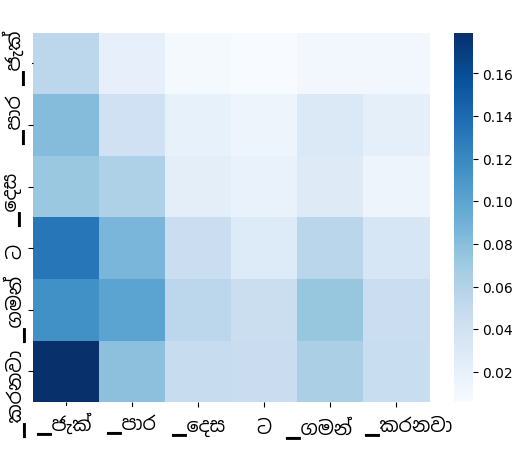}
    \end{minipage}
\end{figure}

From a linguistic perspective, different words in a sentence have different linguistic properties.  Previous work has demonstrated that Pre-trained Language Models (PLMs) capture the notion of syntactic structures and grammatical properties in the language~\cite{nastase2024tracking,nastase2023grammatical,aoyama2022probe}. Named Entities (NEs), Verbs and Nouns significantly contribute to defining the syntactic structure and the semantics of the sentence. Further, these elements play a crucial role in establishing syntactic relationships such as subject-verb agreement, which is generally stronger than those between other words in the sentence. To highlight the prominence of NEs, Verbs and Nouns in a sentence, we visualize the self-attention weight matrix in terms of a heatmap for an English sentence \textit{Jack walks towards the road}, and its Sinhala translation in Figure~\ref{fig:attention_matrix}. In the English sentence, the words "Jack" (NE) and "walk" (Verb) get the highest attention from other words. Similarly, the words which gets the highest attention in Sinhala is a Named Entity and a Noun.

Based on this hypothesis, we introduce a linguistically motivated masking strategy named \textit{Linguistic Entity Masking (LEM)}. In LEM, we limit to masking a single token from the linguistic entity span. As linguistic entities, we consider NEs, nouns and verbs in the sentence. Masking a single token contrasts existing span masking techniques~\cite {sun2019ernie,joshi2020spanbert,levine2020pmi}, which have masked consecutive token spans of the selected n-gram words. We apply LEM on both monolingual ($LEM_{mono}$, akin to MLM) and parallel sentences ($LEM_{para}$, akin to TLM), and conduct continual pre-training on the multiPLM.
We use XLM-R as the multiPLM in our experiments. We continually pre-train XLM-R with $LEM_{mono} + LEM_{para}$ objectives. The resulting model is referred to as XLM-$R_{LEM}$ hereafter. We continually pre-train this same multiPLM with MLM+TLM as well, which serves as our baseline (we term the resulting model XLM-R$_{MLM+TLM}$). We evaluate these models on bitext mining, parallel data curation, as well as code-mixed sentiment classification downstream tasks on three LRL pairs English-Sinhala, English-Tamil, and Sinhala-Tamil. Our results showcase that XLM-$R_{LEM}$  outperforms XLM-R$_{MLM+TLM}$ for these tasks.\\

We carry out extensive experiments (1) to determine the type of monolingual data that is most effective in the first $LEM_{mono}$ continual pre-training step. ie. \textit{dependent monolingual data} (source and target sides of parallel data taken separately as monolingual data) vs \textit{independent monolingual data} (monolingual data without any explicit translation content between the two languages), (2) to empirically evaluate the existing masking strategies (Sub-word masking~\cite{devlin2019bert}, whole-word masking~\cite{devlin2019bert} and span Masking~\cite{joshi2020spanbert}), (3) to identify the most contributing linguistic entity or combination of linguistic entities for masking, (4) to determine the optimal number of tokens for masking within a linguistic entity and (5) to determine the impact of using noisy parallel sentences in the continual pre-training stage.

\noindent Our contributions are as follows:
\begin{itemize}    
    \item We introduce a novel masking strategy, Linguistic Entity Masking (LEM), to improve the cross-lingual representations of existing multiPLMs. We show that XLM-$R_{LEM}$ outperforms XLM-R$_{MLM+TLM}$ for the three downstream tasks, considering English and two LRLs Sinhala and Tamil. 
    \item We conduct an empirical study of the existing masking strategies to evaluate the effectiveness of them on the bitext mining task for LRLs.
    \item We conduct ablation experiments to find the most contributing linguistic entity (among NEs, nouns and verbs) for masking and the number of tokens to mask within the linguistic entity span.
    \item We show that using dependent monolingual corpora than the independent monolingual corpora during the continual pre-training step is favourable for improving the cross-lingual representations. 
\end{itemize}

The rest of the paper is organized as follows. In Section~\ref{sec:related_work} we discuss the related work in the context of pre-training objectives and masking strategies. In Section~\ref{sec:methodology} we describe our methodology, assumptions along with the theoretical justification for the LEM strategy. We describe the experiments in Section~\ref{sec:experiments} and detail out the experimental setup in Section~\ref{sec:exp_setup}. The results are discussed comprehensively in Section~\ref{sec:results_n_discussion}, while additional ablation studies are detailed in Section~\ref{sec:sec_ablation}. The limitations of our approach and potential future directions are highlighted in the Discussion Section~\ref{sec:discussion}, and the paper conclusion is in Section~\ref{sec:conclusion}.

\section{Related Work}\label{sec:related_work} 

\subsection{MLM and TLM Objectives}
Encoder-based multiPLMs such as mBERT and XLM-R were trained on monolingual data using the Masked Language Modeling (MLM) objective. These models have significantly enhanced the performance of various downstream tasks~\citep{rajpurkar2016squad,lai2017race,wang2018glue}.

In BERT (and its multiPLM variant, mBERT), which was trained with MLM\footnote{BERT was also trained using the next sentence prediction task.}, 15\% of the input tokens were randomly selected for corrupting, following a uniform distribution. Out of these, 80\% of the time the tokens were replaced with a [MASK] token, 10\% of the time the tokens were replaced with a random token and 10\% of the time they were left unchanged. Here the MLM objective predicts the corrupted (both masked and replaced) tokens. Successor models such as XLM-R~\citep{conneau2020xlmr} adopt the same 15\% masking percentage and 80\%-10\%-10\% corruption rule during pre-training. The contextualized representations produced by these pre-trained models are then used to obtain sentence embeddings for downstream NLP tasks. However, these models have been reported to be suboptimal for cross-lingual tasks such as bitext mining, 
due to the lack of an explicit objective for improving cross-lingual representations~\citep{hu2021amber}.

To address this limitation,~\citet{conneau2019cross} introduced Translation Language Modeling (TLM), which extended the MLM objective using parallel data. TLM accepts a concatenated pair of parallel sentences as input, and tokens were masked from both sentences. The
rationale was to utilize the context of its translation counterpart to accurately predict the masked token, there by strengthening the cross-lingual capability. TLM was applied in a continual pre-training step, on top of the MLM pre-trained model. In this setting, the MLM step was still required to learn the linguistic information inherent to the languages, while the TLM step strengthened the cross-lingual signal across the language pairs.

\subsection{Different Masking Strategies}
In BERT's MLM strategy, the sub-words were masked. Subsequent work experimented by varying the type of tokens for masking, as summarized in Table~\ref{tab:lr_existing_masking_strategies}. The follow-up work masked consecutive sub-words in text spans~\cite{joshi2020spanbert} and correlated text spans with Point-wise Mutual Information (PMI) masking~\cite{levine2020pmi}.~\citet{zhuang2023heuristic} proposed a heuristic-masking strategy where they considered the unmasked token prediction in addition to the masked token prediction during language model pre-training.~\citet{golchin2023not} utilized an in-domain keyword masking strategy for domain adaptation of the PLM. Most of these techniques have primarily relied on monolingual data and have been evaluated predominantly on high-resource languages.

Closest to our work is Entity/Phrase masking~\citep{sun2019ernie}. This contrasts with our work in three ways. Entity/Phrase masking, selected NEs, noun \textit{phrases} and verb \textit{phrases} for masking. As per analysis in Fig~\ref{fig:attention_matrix}, we considered only verb and noun \textit{words}, in addition to NEs for masking. Secondly, while Entity/Phrase masking masked all consecutive tokens within NEs or Noun/Verb \textit{phrases} identified by a chunking tool, LEM takes a more targeted approach by limiting masking to a single token within the selected linguistic entity span. Finally, they followed a multi-staged pre-training approach. The pre-training stages were, sub-word masking similar to BERT, followed by Phrase masking and Named Entity masking. In comparison, our approach's two continual pre-training stages apply the same LEM strategy with monolingual data and parallel data. They have evaluated the strategy only on High Resource Languages English and Chinese. Further, this strategy has not been extended with parallel data for cross-lingual improvement.

\begin{table*}[h]\centering
\scriptsize
\caption{Existing masking strategies. The \textit{Masked Token Type} indicates the type of words considered for masking.}\label{tab:lr_existing_masking_strategies}
\renewcommand{\arraystretch}{1.35}
\resizebox{\textwidth}{!}{%
\begin{tabular}{lll}\toprule
\textbf{Masking Strategy} &\textbf{Pre-training} &\textbf{Masked token Type} \\\midrule
Sub-word Masking &Pre-training &sub-words \\
Whole-Word Masking &Pre-training &all sub-words in the word \\
Entity/Phrase Masking~\cite{sun2019ernie} &Multi-stage Pre-training &all sub-words in the Named Entity/Noun Phrase \\
Span Masking (spanBERT)~\cite{joshi2020spanbert} &Pre-training &all sub-words in the word n-gram span \\
Point-wise Mutual Information (PMI) Masking~\cite{levine2020pmi} &Pre-training &all sub-words in the correlated word-spans \\
\bottomrule
\end{tabular}}
\end{table*}

\citet{wettig2023should} conducted an empirical study on what tokens should be masked and in what percentages. However, their study was limited to the English language and focused only on downstream tasks such as classification and question-answering. To date, no empirical study has explored these alternative masking strategies specifically for sentence retrieval tasks, particularly in the context of LRL pairs.

\section{Methodology}\label{sec:methodology} 

\begin{figure}[!ht]%
\centering
\includegraphics[width=1.03\textwidth]{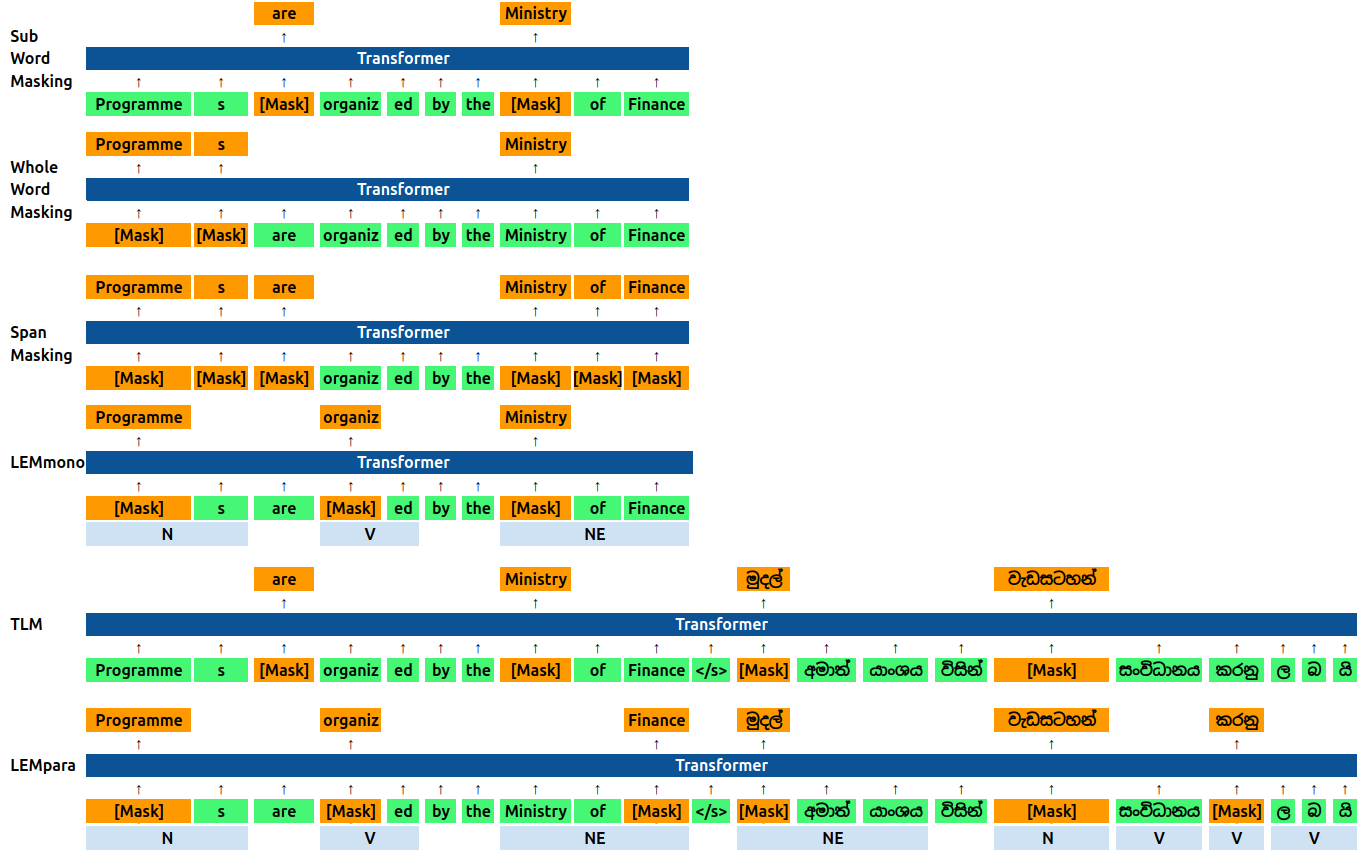}
    \caption{A comparison of the existing masking strategies considering an example from the English-Sinhala language pair. Sub-word masking, Whole Word masking, span masking, and $LEM_{mono}$ consider only monolingual sentences during masking. TLM and $LEM_{para}$ consider concatenated parallel sentences to apply the masking. In $LEM_{mono}$ and $LEM_{para}$, only a single token from the linguistic entity is masked.}\label{fig_methodology}
\end{figure}

In this section, we discuss in detail the LEM strategy. A comparison between our masking strategy and existing masking strategies is shown in Figure~\ref{fig_methodology}. Instead of pre-training a multiPLM from scratch—a computationally expensive process—we leverage LEM in a continual pre-training step. This is a widely adopted approach to improve multiPLMs with respective to representation improvements~\cite{conneau2019cross, feng2022language}.

\begin{figure}[H]%
\centering
\includegraphics[width=0.8\textwidth]{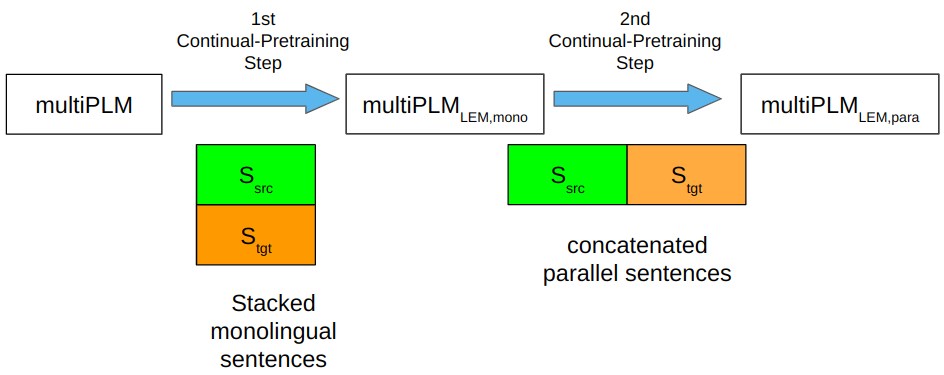}
    \caption{The LEM continual pre-training process. As \textit{multiPLM}, we select an existing multilingual pre-trained language model. 
    The first step ie. LEM$_{mono}$ is to continually pre-train with \textit{stacked monolingual sentences}, meaning the monolingual data from the source side is passed first, followed up by the target language monolingual data. In the second continual pre-training step ie. LEM$_{para}$, the LEM strategy is applied on the \textit{concatenated parallel data}.}\label{fig:continual_training_pipeline}.
\end{figure}

Figure~\ref{fig:continual_training_pipeline} illustrates our two-stage continual pre-training process. Similar to the MLM and TLM training sequence used in XLM~\cite{conneau2019cross}, on top of the multiPLM, we apply the continual pre-training with monolingual and parallel data respectively.

\subsection{Theoretical Framework for Linguistic Entity Masking (LEM)}\label{sec:methodology_LEM}

The theoretical framework of LEM in a monolingual setting ($LEM_{mono}$) can be described as follows: \\
Let the monolingual sequence $X$ be defined as $X=x_1~x_2~x_3~...x_i...~x_n$ where $x_i$ is a word and $n$ is the number of words in the sequence. After tokenization, sequence $X$ can be represented as $\bar{X}$ as in Eq.~\ref{eq:after_tokenization}.

\begin{equation}
\bar{X}=\bar{x}_{1}~\bar{x}_{2}~\bar{x}_{3}~\bar{x}_{4}.....~\bar{x}_{j}....~\bar{x}_{m}\label{eq:after_tokenization}
\end{equation}

\noindent
Here, $~\bar{x}_{j}$ is a token (sub-word) and $m$ is the total number of sub-words returned by the tokenizer. From this sequence, the linguistic entities NEs, verbs and nouns are identified, and $\bar{X}$ can now be represented as a collection of linguistic entities as shown in Eq.~\ref{eq:lem_sampling}. From these linguistic entities, a single token is sampled over a uniform distribution, up to a total of 15\% for masking. If 15\% cannot be obtained from linguistic entities, the remainder would be sampled from the remaining tokens. We use the same corruption rule, 80\%-10\%-10\% as BERT.

\begin{equation}
\bar{X} = \{\{\bar{x}_1~\bar{x}_2\},...  ~\{\bar{x}_4 \bar{x}_5 \bar{x}_6\},.....\{\bar{x}_m\}\}\label{eq:lem_sampling}
\end{equation}
 
During training, the cross-entropy loss ($\mathcal{L}_{LEM_{mono}}$) for masked token prediction, as in Eq.~\ref{eq:ce_loss_mlm} is minimized. In the equation, N is the total number of tokens for prediction and $y_j$ is the expected true label.

\begin{equation}
     \mathcal{L}_{LEM_{mono}} = -\frac{1}{N}\sum_{j=1}^{N} y_j \log(P(x_j))\label{eq:ce_loss_mlm}
\end{equation}

Finally, we extend the TLM objective with LEM into the parallel data setting (LEM$_{para}$). Here we concatenate the source sentence ($X=x_1~x_2~x_3~......~x_m$) and target sentence ($Y=y_1~y_2~y_3~......~y_n$) from the parallel sentence-pair as a single input example and obtain the tokenized output as represented by $\bar{Z}$ in Eq.~\ref{eq:lem_para_after_tokenization}. $\bar{X}=\bar{x}_{1}~\bar{x}_{2}~\bar{x}_{3}.......~\bar{x}_{k}$ and $\bar{Y}=\bar{y}_{1}~\bar{y}_{2}~\bar{y}_{3}.......~\bar{y}_{l}$ are the tokenized source and target sentences, respectively. $k$ and $l$ are the number of tokens (sub-words) in the source and target sentences (respectively) after tokenization.

\begin{equation}
\bar{Z}=\bar{x}_{1}~\bar{x}_{2}~\bar{x}_{3}.......~\bar{x}_{k}~\bar{y}_{1}~\bar{y}_{2}~\bar{y}_{3}.......~\bar{y}_{l}\label{eq:lem_para_after_tokenization}
\end{equation}

Similar to LEM$_{mono}$, in this step, a single token from each linguistic entity (NE, verb or noun) from both languages are selected for corruption according to the 80\%-10\%-10\% rule. If 15\% of linguistic units were not found in the sequence, the balance is sampled from the remaining tokens. During training, the corrupted token prediction cross-entropy loss, ($\mathcal{L}_{LEM_{para}}$) (Eq.~\ref{eq:ce_tlm_loss}) is minimized. S and T correspond to the total number of tokens masked from the source and target side sentences respectively. $z_s$ and $z_t$ are the true tokens to be predicted.

\begin{equation}
     \mathcal{L}_{LEM_{para}} = -\frac{1}{S}\sum_{s=1}^{S} z_s \log(P(x_s)-\frac{1}{T}\sum_{t=1}^{T} z_t \log(P(j_t))\label{eq:ce_tlm_loss}
\end{equation}

Languages such as Sinhala and Tamil exhibit morphological richness, requiring words to be inflected based on attributes such as number, gender, and case category. Table~\ref{tab:word_inflections_examples} shows examples for such word inflections. Additionally, the presence of out-of-vocabulary words in LRLs often leads to an increased number of sub-words after tokenization. Therefore, approaches like whole-word masking, span masking, or entity/phrase masking tend to mask longer spans. This reduction in context weakens the ability to accurately predict the masked tokens, ultimately hindering representation learning. In contrast, LEM mitigates this issue by masking a single token from a linguistic entity, which we empirically prove in Section~\ref{subsec:ablation_noof_tokens}.

\begin{table}[!htb]
    \centering
    \caption{English (En), Sinhala (Si) and Tamil (Ta) examples of the returned sub-words after the tokenization step. English nouns get inflected based on the number only. But Sinhala and Tamil nouns get inflected based on case and gender as well.}\label{tab:word_inflections_examples}
    \begin{tabular}{c}
    \includegraphics[width=\linewidth]{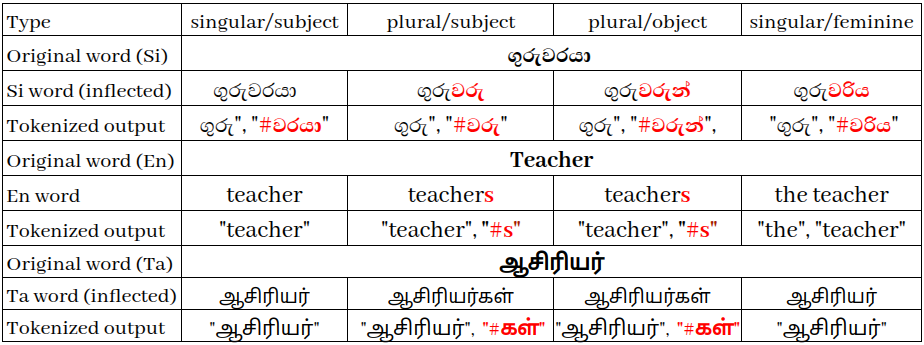} \\
    \end{tabular}
\end{table}

\section{Experiments}\label{sec:experiments} 

\subsection{Impact of the type of monolingual data in $LEM_{mono}$ }\label{sec:exp_mono_vs_parallel_ablation}
We experiment with independent and dependent monolingual data to observe its impact on the continual pre-training step $LEM_{mono}$.
We sample 60,000 sentences from the MADLAD-400 and all the available 60,000 sentences from SiTa-Trilingual dataset for each language. Next, we continually pre-train XLM-R with those datasets separately with $LEM_{mono}$ and evaluate on the bitext evaluation dataset. 

To assess whether increasing the independent training data would yield in any improvements, we repeat the experiment with a sample size of 100,000 from MADLAD-400. Finally, as an extreme case, we conduct a third experiment using 500,000 sentences for the Sinhala-Tamil language pair. Here, 60,000, 100,000, and 500,000 correspond to the training data size per language, with the full training set size being double the amount specified. We conduct this evaluation for all three language pairs.

\subsection{Evaluation of Different Masking Strategies}\label{sec:diff_masking_strategies}
We empirically evaluate various masking strategies and assess their performance on the bitext mining task. The masking strategies explored in this study are as follows: \\

\noindent\textbf{Sub-word Masking} - Following the BERT MLM, with each sentence, 15\% of tokens are selected randomly and corrupted according to 80\%-10\%-10\% rule. \\
\noindent\textbf{Whole Word Masking} - All the sub-words corresponding to the randomly sampled words are masked. A total of 15\% tokens are sampled and corrupted according to 80\%-10\%-10\% rule. \\
\textbf{Span Masking} - Consecutive word spans are sampled over a geometrical distribution and 15\% of tokens are masked. The masking is limited to whole-word tokens as defined in the original work.

\subsection{Evaluation of LEM Strategy and Ablation Study}\label{sec:exp_lem}
This section describes the ablation experiments we conduct to determine the most contributing linguistic entity or their combination in the LEM strategy. We use the baselines as described in Section~\ref{subsec:baselines}. 

We identify the NEs in English, Sinhala, and Tamil sentences, using an in-house fine-tuned multilingual NER model~\cite{ranathunga2024multi}. To identify nouns and verbs in the sequences, we employ Flair POS tagger~\citep{akbik2018coling} for English, the Sinhala TnT POS Tagger~\citep{fernando2018evaluation,fernando2016comprehensive} for Sinhala, and ThamizhiUDp~\citep{sarveswaran2020thamizhiudp} for Tamil. Flair reported an F1 score of 98.19\% and is the best model for English POS Tagging. The Sinhala POS  Tagger had been trained using SVM and has an overall accuracy of 84.68\% with a 59.86\% accuracy for tagging unknown words. The Tamil POS Tagger is a neural-based model, with a F1 score of  93.27\%. For Sinhala and Tamil, these are the models that returned optimal results.

The initial ablation experiment masks a single linguistic entity type, such as only NEs, only verbs, or only nouns. Subsequently, combinations of these linguistic entity types are examined.

In our experiments, 100\%NE+15\%MLM means, priority is given for sampling from NEs. If it does not produce enough tokens for masking, then the balance is sampled from the remaining tokens. When combining several linguistic entities, e.g.~100\%NE+100\%VB+15\%MLM means, priority is given to sample the tokens for masking from both NEs and verbs.

\subsection{Evaluation Tasks}
We evaluate the success of our LEM masking strategy on three downstream tasks - bitext mining, parallel data curation and code-mixed sentiment classification. 

\subsubsection{Bitext Mining}\label{sec:exp_eval_bitext} 

Bitext mining is a sentence retrieval task that retrieves a target language translation for a given source sentence or vice versa from a document-aligned dataset. The performance of bitext mining relies heavily on the quality of cross-lingual embeddings. Recent bitext mining techniques are embedding-based, where they identify the translation pairs based on the semantic distance between the source sentences and candidate target sentences. Improvements in bitext mining techniques can be categorized into two: (1) refining the semantic similarity distance calculation function~\citep{artetxeschwenk2019margin,fernando2023exploiting} between sentence embeddings and (2) enhancing cross-lingual sentence representations \citep{artetxe2019massively,yang2020multilingual,feng2022language}. Our LEM strategy aims to improve the latter.\\

We obtain the sentence embeddings from XLM-R$_{LEM}$ as well as from the baselines, and use margin-based cosine similarity~\citep{artetxeschwenk2019margin} function to identify parallel sentence pairs. We choose margin-based cosine similarity over conventional cosine similarity for this task due to its lower rate of false positives. Then we rank the parallel sentences according to their similarity scores. Bitext mining is performed using the three criteria: Forward (FW), Backward (BW), and Intersection (IN)~\citep{artetxeschwenk2019margin}. FW retrieves the target sentence for each source sentence, BW retrieves the source sentence for each target sentence, and IN considers the intersection of the parallel sentences retrieved using FW and BW criteria.
The Recall evaluation metric is used to report the bitext mining results.
 
\subsubsection{Parallel Data Curation}\label{sec:exp_data_curation} 

\noindent Although large-scale bitext mining~\citep{schwenk2021ccmatrix,costa2022nllb} alleviates the parallel data scarcity problem in NMT, they are mostly noisy~\citep{kreutzeretal2022quality,ranathunga2024quality}. Therefore, a parallel data curation step is crucial to filter out noisy parallel sentences from the corpus. This is done by obtaining sentence representations from a multiPLM and by calculating the semantic similarity using cosine distance~\cite {feng2022language,costa2022nllb} between each parallel sentence pair. Then the parallel sentences are sorted in descending order and the top-most ranked parallel sentences are used to train the NMT system.

To further evaluate the effectiveness of these models with improved cross-lingual representations, we conduct a parallel corpus curation task and perform an extrinsic evaluation by training NMT systems with the top-ranked sentences.

First, we rank the parallel sentences for translation quality using the baseline (Section~\ref{subsec:baselines}) and the XLM-R$_{LEM}$ models. Using the top 50,000 sentences from the ranked parallel sentence pairs, we train NMT systems for each language pair. NMT scores are reported on the Flores+ devtest, using sacreBLEU~\citep{post2018sacrebleu}, ChrF~\citep{popovic2015chrf}, ChrF++~\cite{popovic2017chrfpp} and spBLEU~\citep{goyal2022flores} metrics.
We base the discussion of the results using the ChrF metric while we include the full results Table~\ref{tab:pdc_nmt_scores} in Appendix~\ref{secB}.

\subsubsection{Code-Mixed Sentiment Analysis} \label{sec:exp_code_mixed_sent_analysis}

MultiPLMs fine-tuned on task-specific monolingual datasets has achieved the state-of-the-art performance in sentiment analysis tasks~\cite{barbieri2022xlm}. However, sentiment analysis on code-mixed data remains a challenging task. Code-mixing~\cite{myers1993duelling} occurs when linguistic units—such as phrases, words, or morphemes—from one language are embedded into the utterance or sentence of another language. In this setting, the performance of sentiment analysis on code-mixed data largely depends on the quality of the cross-lingual embeddings learned by the MultiPLM.

We conduct the baseline experiments (Section~\ref{subsec:baselines}) and compare them with the results we obtain with the XLM-R$_{LEM}$ model. Here we use each encoder model by fine-tuning them on an English-Sinhala code-mixed task. We follow a two-step fine-tuning, where the first fine-tuning is done using the English Amazon product review sentiment analysis dataset. Here we report the zero-shot scores on the code-mixed EnSi evaluation set. Finally, the intermediate model is further fine-tuned on the English-Sinhala code-mixed dataset and the results are reported. We use Precision, Recall and F1 to report the evaluation scores for the sentiment analysis task.

\section{Experiment Setup}\label{sec:exp_setup} 
\subsection{Data Selection}\label{sec:exp_data_selection}

We consider English, Sinhala and Tamil for our experiments. Sinhala and Tamil are the official languages of Sri Lanka and English is used as a link language. They belong to three distinct language families; Indo-European, Indo-Aryan and Dravidian language families, respectively. Further, Sinhala and Tamil are low-resource and medium-resource languages respectively~\citep{joshi2020state, ranathunga2022some}. They are also morphologically rich languages. Sinhala, in particular, is only used in the island nation of Sri Lanka and has seen slow progress in language technologies~\citep{ranathunga2022some,de2023survey}.\\

\noindent\textbf{Monolingual and Parallel Data:} As elaborated in Section~\ref{sec:exp_mono_vs_parallel_ablation}, we carry out an ablation study to determine the type of monolingual data that is most suitable for the first continual pre-training step. We obtain the independent monolingual data from MADLAD-400~\citep{kudugunta2024madlad}. It is a collection of document-level data of 3 Trillion tokens from  Common Crawl\footnote{https://commoncrawl.org/} for 419 languages. As the dependent-monolingual data, we obtain the monolingual sides from the SiTa-Trilingual parallel dataset~\cite{fernando2020data}. It is a human-curated gold standard three-way parallel dataset between Sinhala-Tamil-English languages with 60,000 training data.

We preprocess the MADLAD-400 data to extract clean sentences for each language as follows: First, the document-level data is segmented into sentences using the nltk\footnote{https://www.nltk.org/index.html} sentence tokenizer. We then filter these sentences using the LID (\textbf{L}anguage \textbf{ID}entification) model\footnote{https://github.com/gordicaleksa/Open-NLLB}. Subsequently, we remove noisy data, including HTML tags, URLs, and sentences with less than 60\% textual content. Finally, religious texts were excluded through a keyword filter.\footnote{Keywords being bible book names along with common words from the bible.} However, for SiTa-Trilingual data, no preprocessing was applied as the data was of high quality. \\

\noindent\textbf{NLLB/CCAligned Datasets:} For the parallel data curation task, we obtain parallel data from NLLB~\citep{costa2022nllb} and CCAligned~\citep{el2020ccaligned} corpora. Both these corpora provide parallel data for the three language pairs: English-Sinhala,  English-Tamil, and Sinhala-Tamil. NLLB and CCAligned are known to be noisy parallel data for the considered language pairs~\cite{ranathunga2024quality}.\\

\noindent\textbf{ParaCrawl Dataset:} To analyse the performance of LEM strategy with noisy data, we select the English-Sinhala ParaCrawl~\cite{banon2020paracrawl} dataset. It is a web-mined parallel corpus with 217,412 parallel sentences.\\

\noindent\textbf{Code-mixed Sentiment Classification pairsDataset:} For the code-mixed sentiment classification task, we use the English-Sinhala code-mixed dataset~\cite{rathnayake2022adapter} of 13,521 sentences. We have experimented with the English-Sinhala language pair.\\

\noindent\textbf{Amazon Product Review Dataset:}
To report the zero-shot scores for the English-Sinhala sentiment classification task, we used the Amazon product review sentiment analysis dataset~\footnote{https://www.kaggle.com/datafiniti/consumer-reviews-of-amazon-products}. The full dataset has training data of 3.6M and test data 400,000 respectively. During fine-tuning, we sample only 100,000 as training data, in order to avoid catastrophic forgetting~\cite{koloski2023measuring} of the cross-lingual representations in XLM-$R_{LEM}$\\

\noindent\textbf{Trilingual Bitext Mining Evaluation Set:} For the bitext mining task, we use an existing human-created dataset ~\citep{fernando2023exploiting}. It consists of trilingual data obtained from four Sri Lankan news sources Army\footnote{https://www.army.lk/}, Hiru\footnote{https://www.hirunews.lk/}, ITN\footnote{https://www.itnnews.lk/} and Newsfirst\footnote{https://english.newsfirst.lk/}. For each news source, there are human-aligned 300 sentence-pairs.\\

\noindent\textbf{Flores+ Evaluation Set:} For the NMT experiments, we use dev and devtest splits from Flores+\footnote{https://github.com/openlanguagedata/flores} as the validation and evaluation sets, respectively.

\subsection{multiPLM Selection}

We select XLM-R as the base multiPLM for our experiments. Other popular multiPLMs, XLM and mBERT do not cover Sinhala language. XLM-R has already demonstrated promising performance in downstream tasks for the low-resource languages considered in this study~\citep{dhananjaya2022bertifying, rathnayake2022adapter, udawatta2024use, ranathunga2024multi}. It is a 278M parameter model, pre-trained for 100 languages. However, the amount of Sinhala and Tamil data used during XLM-R pre-training is much lower than that for English (English 55B, Sinhala 243M, Tamil 595M).

\subsection{Baselines}\label{subsec:baselines}
In our evaluation of downstream tasks, we set up two baseline experiments. 
\begin{itemize}
    \item \textbf{XLM-R} - Obtain embeddings from the out-of-the-box XLM-R pre-trained model. 
    \item \textbf{XLM-R\textbf{$_{MLM+TLM}$}}~\cite{conneau2019cross} - We continually pre-train the XLM-R with MLM+TLM objectives and use the representation improved encoder to obtain embeddings.
\end{itemize}

\subsection{Implementation and Hyper-parameters}

\subsubsection{Linguistic Entity Masking (LEM)} 
We customize the MLM training implementation released with the sentence-transformers\footnote{https://www.sbert.net/} library (built on Huggingface transformers\footnote{https://huggingface.co/docs/transformers/index}), to support XLM-R tokenization and implement the LEM strategy. Each continual pre-training experiment is executed for 60 epochs with early stopping. Then the checkpoint with the least validation loss is selected as the best-performing model. The experiments are conducted on Nvidia Quadro RTX 6000 GPU with 24GB VRAM. The hyper-parameters of XLM-R~\footnote{https://huggingface.co/FacebookAI/xlm-roberta-base} model and other training parameters used in the continual pre-training experiments are shown in Table~\ref{tab:exp_hyperparameters}.

\begin{table}[h]
\caption{Hyper-parameters used during continual pertaining with LEM strategy}\label{tab:exp_hyperparameters}
\renewcommand{\arraystretch}{1.2}
\begin{tabular*}{0.8\textwidth}{@{\extracolsep\fill}lr}
\toprule
\textbf{Hyperparameter} &\textbf{Argument value} \\
\midrule
No of Layers &12 \\
Hidden Size &768 \\
Attention Heads &12 \\
hidden\_dropout\_prob &0.1 \\
Learning Rate &5e-3 \\
Training batch-size &32 \\
Sequence Length &120 \\
Adam $\epsilon$ &1 e-08 \\
Adam $\beta_1$ &0.9 \\
Adam $\beta_2$ &0.99 \\
\midrule
\end{tabular*}
\end{table}

\subsubsection{NMT Experiments}
We obtain the top-ranked sentences for the NMT experiments and train a Sentencepiece\footnote{https://github.com/google/sentencepiece} tokenizer with a vocabulary size of 25000. Then we use fairseq toolkit ~\citep{ott2019fairseq} to model and train the vanilla transformer-based Sequence-to-Sequence NMT model. The experiments are conducted on a Nvidia Quadro RTX6000 GPU with 24GB VRAM. The hyper-parameters used during training along with the training parameters are shown in Table~\ref{tab:nmt_hyperparameters}. Each experiment is run for 100 epochs with the early stopping criteria.

\begin{table}[htp]\centering
\caption{Training parameters for NMT experiments.}\label{tab:nmt_hyperparameters}
\scriptsize
\renewcommand{\arraystretch}{1.2}
\begin{tabular*}{0.8\textwidth}{@{\extracolsep\fill}lr}
\toprule
\textbf{Hyper-parameter} &\textbf{Argument value} \\
\midrule
encoder/decoder Layers &6 \\
encoder/decoder attention heads &4\\
encoder-embed-dim &512\\
decoder-embed-dim &512\\
encoder-ffn-embed-dim &2048 \\
decoder-ffn-embed-dim &2048 \\
dropout &0.4\\
attention-dropout &0.2\\
optimizer &adam\\
Adam $\beta_1$, Adam $\beta_2$ &0.9, 0.99 \\
warmup-updates &4000\\
warmup-init-lr &1e-7\\
learning rate &1e-3\\
batch-size &32\\
patience &6\\
fp16 &True\\
\midrule
\end{tabular*}
\end{table}

\subsubsection{Code-Mixed Sentiment Analysis}
For the sentiment classification experiments, we fine-tune both the baseline models and the XLM-R$_{LEM}$ model. A linear layer is added as the classification head to facilitate binary sentiment classification. Full fine-tuning is performed, allowing updates to all model parameters, including those in the classification head, during training. The Hyper-parameters used during these experiments are presented in Table~\ref{tab:sent_class_hyperparameters}. We train for 20 epochs with early stopping.

\begin{table}[h]\centering
\caption{Training parameters for the sentiment classification task. experiments.}\label{tab:sent_class_hyperparameters}
\scriptsize
\renewcommand{\arraystretch}{1.2}
\begin{tabular*}{0.8\textwidth}{@{\extracolsep\fill}lr}
\toprule
\textbf{Hyper-parameter} & \textbf{Argument value} \\
\hline
Number of Layers & 12\\
Hidden Size & 768 \\
Attention Heads & 12 \\
Hidden dropout prob & 0.1 \\
Learning Rate & 2e-5 \\
Weight Decay  & 0.01 \\
Training Batch Size & 128\\
Sequence Length & 80\\
Adam $\epsilon$ & 1e-08 \\
Adam $\beta_1$ & 0.9 \\
Adam $\beta_2$ & 0.99\\ 
\hline
\end{tabular*}
\end{table}

\subsubsection{Improving Continual Pre-training Efficiency}
Named Entity Recognition (NER) and Part-of-Speech (POS) tagging during training increases training time drastically. We introduce a pre-processing step to mitigate this issue. Specifically, a dictionary is created to store the linguistic entities, such as named entities, verbs, and nouns, for each sentence. We maintain a sub-word-level mapping in the dictionary, allowing for precise token identification while reducing computational overhead during training.

\section{Results and Discussion}\label{sec:results_n_discussion}

\subsection{Impact of the type of monolingual data in $LEM_{mono}$}

\begin{figure}[!ht]%
\centering
\includegraphics[width=1.03\textwidth]{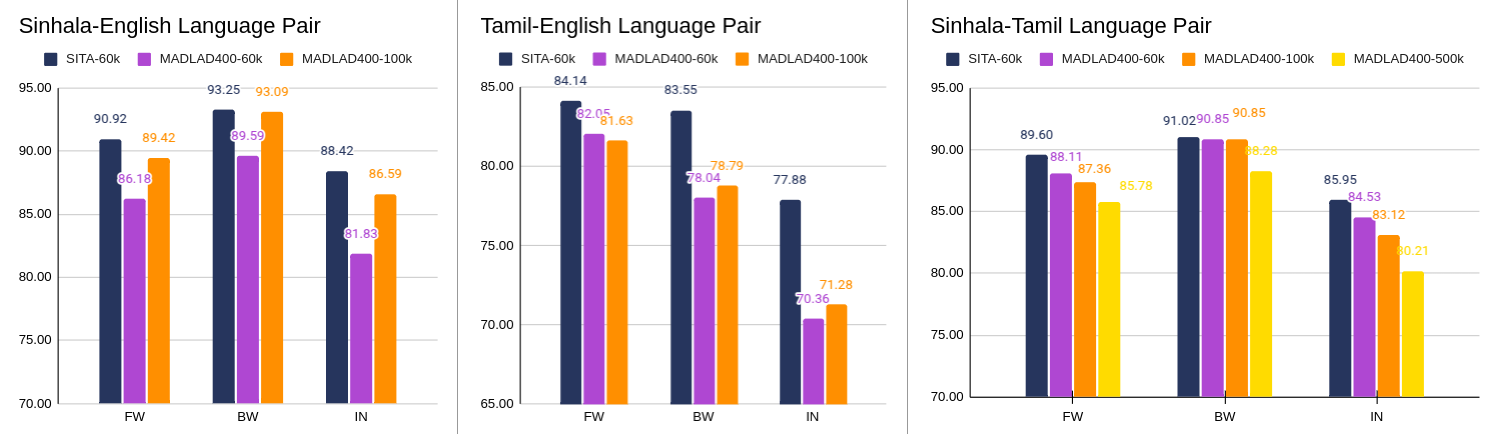}
    \caption{Bitext mining Recall scores for using independent monolingual data (MADLAD-400) versus dependent monolingual data obtained from the parallel corpus (SiTa-Trilingual parallel Corpus).}\label{fig:mono_vs_parallel_mlm}
\end{figure}

Figure~\ref{fig:mono_vs_parallel_mlm} shows the bitext mining results for the $LEM_{mono}$ step using independent and dependent monolingual data. The detailed results are available in Table~\ref{tab:mono_vs_parallel_mlm} in Appendix ~\ref{secA}.

The highest performance was observed for the dependent monolingual data, a trend consistent across all three language pairs. Surprisingly, when the independent set was increased to 100,000, this increase did not surpass the scores obtained using the dependent monolingual 50,000 training dataset. This was evident in all three language pairs. When the dataset size was further increased to 500,000, the results further deteriorated.\footnote{Due to resource constraints and the reduced performance for the Sinhala-Tamil direction, the experiment using 500,000 was not conducted for the other two language pairs.}
Therefore, it is evident that utilizing dependent monolingual data is advantageous during the LEM$_{mono}$ step to enhance cross-lingual representations.

\subsection{Evaluation of Different Masking Strategies}
Table~\ref{tab:mlm_stratergies} shows the experimental results for the bitext mining task using XLM-R continually pre-trained with different MLM strategies (Section~\ref{sec:diff_masking_strategies}). Based on the averaged recall scores, the baseline XLM-R consistently delivered the highest performance in bitext mining tasks. An exception was observed for the Sinhala-Tamil BW criterion, where the sub-word masking strategy outperformed the baseline. However, a key observation was that continual pre-training with the existing masking strategies in general deteriorated the already learnt cross-lingual representations in the XLM-R model.

As outlined in Section~\ref{sec:methodology_LEM}, morphologically rich, low-resource languages tend to generate a higher number of sub-word tokens during the tokenization process due to the presence of infrequent words in the sequences. Consequently, masking strategies like whole-word masking and span masking result in longer masked spans, which reduce the available context for accurate predictions. We hypothesize that this reduced context reduces the prediction accuracy, thus weakening the already learnt representations in the XLM-R model. As a result, the existing masking strategies return degraded results.

\begin{table*}[h]\centering
\caption{Bitext mining Recall scores for the different masking strategies.}\label{tab:mlm_stratergies}
\scriptsize
\renewcommand{\arraystretch}{1.35}
\resizebox{\textwidth}{!}{%
\begin{tabular}{lrrrrrrrrrrrrrrrr}\toprule
\multirow{2}{*}{\textbf{Experiment}} &\multicolumn{3}{c}{\textbf{Army}} &\multicolumn{3}{c}{\textbf{Hiru}} &\multicolumn{3}{c}{\textbf{ITN}} &\multicolumn{3}{c}{\textbf{Newsfirst}} &\multicolumn{3}{c}{\textbf{Averages}} \\\cmidrule{2-16}
&\textbf{FW} &\textbf{BW} &\textbf{IN} &\textbf{FW} &\textbf{BW} &\textbf{IN} &\textbf{FW} &\textbf{BW} &\textbf{IN} &\textbf{FW} &\textbf{BW} &\textbf{IN} &\textbf{FW} &\textbf{BW} &\textbf{IN} \\\midrule
\multicolumn{16}{c}{\textbf{English - Sinhala }} \\
XLM-R &\textbf{92.33} &93.33 &\textbf{89.67} &\textbf{96.35} &\textbf{96.68} &\textbf{95.68} &\textbf{94.00} &96.00 &92.33 &\textbf{96.67} &\textbf{95.33} &\textbf{94.33} &\textbf{94.84} &\textbf{95.34} &\textbf{93.00} \\
Sub-word Masking &88.33 &\textbf{93.67} &85.33 &92.03 &93.36 &89.70 &91.67 &\textbf{96.67} &\textbf{93.67} &91.67 &95.33 &90.00 &90.92 &94.76 &89.68 \\
Whole-word Masking &87.33 &92.67 &85.33 &95.02 &94.01 &94.02 &93.00 &91.67 &90.33 &93.67 &93.67 &91.67 &92.25 &93.00 &90.34 \\
Span Masking &89.00 &89.67 &85.00 &95.02 &94.02 &92.03 &90.33 &91.67 &85.67 &93.67 &92.67 &90.33 &92.00 &92.01 &88.26 \\
\hline
\multicolumn{16}{c}{\textbf{English - Tamil }} \\
XLM-R &\textbf{86.67} &\textbf{88.33} &\textbf{82.00} &\textbf{83.00} &\textbf{78.33} &\textbf{72.67} &83.22 &\textbf{83.56} &\textbf{78.86} &\textbf{92.33} &\textbf{91.33} &89.33 &\textbf{86.31} &\textbf{85.39} &\textbf{80.71} \\
Sub-word Masking &84.00 &86.00 &77.67 &80.33 &75.00 &68.33 &\textbf{83.56} &82.21 &78.52 &90.67 &91.00 &\textbf{89.67} &84.64 &83.55 &78.55 \\
Whole-word Masking &83.33 &87.33 &77.67 &78.67 &73.33 &64.33 &80.20 &80.87 &75.84 &85.67 &91.00 &83.67 &81.97 &83.13 &75.38 \\
Span Masking &82.67 &83.00 &75.33 &78.67 &76.67 &69.33 &83.22 &82.22 &76.85 &89.67 &90.00 &85.67 &83.56 &82.97 &76.79 \\
\hline
\multicolumn{16}{c}{\textbf{Sinhala-Tamil}} \\
XLM-R &83.44 &81.46 &78.15 &\textbf{90.67} &91.00 &\textbf{87.33} &\textbf{91.33} &90.00 &87.00 &\textbf{93.67} &\textbf{95.33} &\textbf{92.33} &\textbf{89.78} &89.45 &\textbf{86.20} \\
Sub-word Masking &\textbf{86.75} &88.08 &\textbf{81.96} &88.00 &89.33 &84.00 &93.33 &\textbf{92.67} &\textbf{89.33} &90.33 &94.00 &89.00 &89.60 &\textbf{91.02} &86.07 \\
Whole-word Masking &85.76 &\textbf{89.73} &81.46 &88.33 &\textbf{91.33} &84.67 &90.33 &90.33 &86.67 &90.00 &91.67 &87.67 &88.61 &90.77 &85.11 \\
spanMasking &85.78 &85.10 &81.79 &88.67 &91.00 &87.00 &91.00 &91.00 &87.33 &89.00 &90.67 &84.33 &88.61 &89.44 &85.11 \\
\bottomrule
\end{tabular}}
\end{table*}

\subsection{Evaluation of LEM Strategy and Ablation Study}\label{sec:results_lem}

The ablation study results for considering each linguistic entity along with combinations of them, on the LEM strategy are available in Tables~\ref{tab:sien_ablation},~\ref{tab:taen_ablation},~\ref{tab:sita_ablation} in the Appendix~\ref{secC}. In Table~\ref{tab:sa_summary}, we summarise the final gains.

For the Sinhala-Tamil language pair, compared to the scores obtained with XLM-R raw embeddings, XLM-R$_{LEM}$ returned the highest gain of +3.1 Recall points. Compared with XLM-R$_{MLM+TLM}$, this was a gain of +1.4 points. For the English-Tamil language pair, the LEM$_{mono}$ step produced a reduced score compared to the XLM-R baseline. However, after the second continual pre-training step XLM-R$_{LEM}$ scores surpassed the XLM-R baseline by +1.2. Further, the highest gain produced by our method compared to XLM-R$_{MLM+TLM}$ was for the English-Tamil language pair, which was +2.4 points. With the English-Sinhala language pair, a similar behaviour was observed. Here XLM-R$_{LEM}$ compared to baseline XLM-R has a gain of +0.4 points while the improvement compared to XLM-R$_{MLM+TLM}$ was +1.7 scores. 

In all these language pairs, the best scores were produced when the linguistic entity NEs were included in the LEM strategy. Due to the consistent gains across the three language pairs, we can safely conclude that LEM is favourable to improving the cross-lingual representations in existing multiPLMs.

\begin{table}[!htp]\centering
\scriptsize
\caption{Results for bitext mining task in terms of recall points. For comparison purposes, the FW, BW and IN gains are averaged and reported in the \textit{Overall Average Gain column}.}\label{tab:sa_summary}
\renewcommand{\arraystretch}{1.35}
\begin{tabular}{lccccc}\toprule
\multirow{2}{*}{\textbf{}} &\multicolumn{3}{c}{\textbf{Average Gains}} &\multirow{2}{*}{\textbf{Overall Average}} \\\cmidrule{2-4}
&\textbf{FW} &\textbf{BW} &\textbf{IN} & \textbf{Gain}\\
\midrule
\multicolumn{5}{l}{\textbf{Sinhala-Tamil}} \\
\midrule
XLM-R$_{LEM}$ vs XLM-R &+2.36 &+4.14 &+2.90 &+3.1 \\
XLM-R$_{LEM}$ vs XLM-R$_{MLM+TLM}$ &+1.95 &+0.48 &+1.83 &+1.4 \\
\midrule
\multicolumn{5}{l}{\textbf{English-Tamil}} \\
\midrule
XLM-R$_{LEM}$ vs XLM-R &+0.75 &+1.59 &+1.17 &+1.2 \\
XLM-R$_{LEM}$ vs XLM-R$_{MLM+TLM}$ &+2.34 &+1.84 &+2.92 &+2.4 \\
\midrule
\multicolumn{5}{l}{\textbf{English-Sinhala}} \\
\midrule
XLM-R$_{LEM}$ vs XLM-R &+0.25 &+0.50 &+0.42 &+0.4 \\
XLM-R$_{LEM}$ vs XLM-R$_{MLM+TLM}$ &+1.50 &+1.50 &+2.08 &+1.7 \\
\bottomrule
\end{tabular}
\end{table}

\subsection{Parallel Data Curation}\label{sec:results_nmt_evaluation} 

Table~\ref{tab:nmt_scores_chrfpp} shows the NMT results for training the top 50,000 sentence pairs obtained by ranking the parallel sentences with the baseline and XLM-R$_{LEM}$ models for the NLLB and CCAligned corpora. It can be observed consistently that both XLM-R$_{MLM+TLM}$ and XLM-R$_{LEM}$ improved models outperform the baseline XLM-R scores.

The NMT results show that the XLM-R$_{LEM}$ model produce superior results compared to XLM-R$_{MLM+TLM}$, for all three language pairs across the two corpora. We believe the magnitude of the gain is dependent on the characteristics of the parallel corpus and the size of the training data sample. For the English-Tamil language pair, the CCAligned corpus produce a significant gain for XLM-R$_{LEM}$ compared to XLM-R$_{MLM+TLM}$. This justifies the effectiveness of the LEM strategy which was not evident with random masking followed in MLM+TLM. The rest of the gains vary from +0.3 to +0.8 ChrF points. According to metric analysis by~\citet{kocmi-etal-2024-navigating}, these gains are equivalent to +0.48 to +1.12 BLEU points with a human accuracy of 54.2\% to 66\% respectively. This means the improvement in the translation quality in the NMT systems is almost in line with a minimum human accuracy rating of 54.2\% to 66\%.

The observations are consistent with other metrics, as shown in Table~\ref{tab:pdc_nmt_scores} in the Appendix~\ref{secB} as well. The results further prove that the scoring from the XLM-R$_{LEM}$ model has managed to identify quality sentence pairs more than the other models. Therefore, improvement in the cross-lingual representations with the LEM strategy benefits the parallel data curation task as well.

\begin{table*}[!htp]\centering
\caption{ChrF scores for the parallel data curation task. The scores have been reported on the Flores+ devtest. The values in brackets indicate the gains of XLM-R$_{LEM}$ compared to the XLM-R and the XLM-R$_{MLM+TLM}$ respectively.}\label{tab:nmt_scores_chrfpp}
\scriptsize
\renewcommand{\arraystretch}{1.35}
\resizebox{0.8\linewidth}{!}{%
\begin{tabular}{lrrrr}\toprule
\textbf{} &\textbf{Sinhala - Tamil} &\textbf{English - Sinhala} &\textbf{English-Tamil} \\
\hline
\multicolumn{4}{c}{\textbf{NLLB}} \\
\hline
XLM-R &38.6 &33.1 &44.00 \\
XLM-R$_{MLM+TLM}$ &41.3 &43.2 &50.70 \\
XLM-R$_{LEM}$ &\textbf{\textcolor{teal}{(+3.5/+0.8)} 42.1} &\textbf{\textcolor{teal}{(+10.8/+0.7)} 43.9} &\textbf{\textcolor{teal}{(+7.2/+0.5)} 51.2} \\
\hline
\multicolumn{4}{c}{\textbf{CCAligned}} \\
\hline
XLM-R &37.2 &10.2 &5.2 \\
XLM-R$_{MLM+TLM}$ &42.3 &33.9 &31.5 \\
XLM-R$_{LEM}$ &\textbf{\textcolor{teal}{(+5.2/+0.3)} 42.6} &\textbf{\textcolor{teal}{(+24.3/+0.6)} 34.5} &\textbf{\textcolor{teal}{(+29.1/+2.8)} 34.3} \\
\bottomrule
\end{tabular}}
\end{table*}

\subsection{Code-Mixed Sentiment Analysis}\label{sub:code_mixed_sa}

Table~\ref{tab:results_code_mixed} summarizes the evaluation scores for the code-mixed sentiment analysis task on the English-Sinhala dataset. The XLM-R$_{LEM}$ model consistently outperformed both XLM-R and XLM-R$_{MLM+TLM}$ when fine-tuned directly on the code-mixed dataset.

In the zero-shot evaluation (fine-tuned on English-only data), all models showed reduced F1 scores. We suspect that the English monolingual fine-tuning step adversely impacted the cross-lingual alignment between the two languages.

When further fine-tuned on the code-mixed dataset, XLM-R$_{LEM}$ achieved the highest scores, indicating that the cross-lingual enhancement benefits the code-mixed sentiment classification task. These results highlight the promise of LEM in improving multilingual models, particularly for low-resource and code-mixed language applications.

\begin{table*}[!h]
\centering
\caption{Scores for the sentiment analysis task in terms of Precision, Recall and F1. The results are reported on the English-Sinhala code-mixed evaluation set.}\label{tab:results_code_mixed}
\scriptsize
\renewcommand{\arraystretch}{1.45}
\resizebox{\linewidth}{!}{%
\begin{tabularx}{\linewidth}{l>{\centering\arraybackslash}X>{\centering\arraybackslash}X>{\centering\arraybackslash}X}
\hline
& \textbf{Precision} & \textbf{Recall} & \textbf{F1} \\
\hline
\multicolumn{4}{l}{\textbf{Fine-tuning on English-Sinhala Code-mixed Sentiment Analysis dataset}} \\
\hline
XLM-R & 88.72\% & 88.72\% & 88.72\% \\
XLM-R$_{MLM+TLM}$ & 88.88\% & 88.88\% & 88.88\% \\
XLM-R$_{LEM}$ & \textbf{89.09\%} & \textbf{89.33\%} & \textbf{89.20\%} \\
\hline
\multicolumn{4}{l}{\textbf{Fine-tuning on English Sentiment Analysis dataset (Zero-shot scores)}} \\
\hline
XLM-R & 74.19\% & \textbf{82.14\%} & \textbf{75.17\%} \\
XLM-R$_{MLM+TLM}$ & \textbf{74.74\%} & 79.57\% & 73.92\% \\
XLM-R$_{LEM}$ & 69.18\% & 76.36\% & 68.13\% \\
\hline
\multicolumn{4}{l}{\textbf{Fine-tuning on English-Sinhala Code-mixed Sentiment Analysis dataset}} \\
\hline
XLM-R & 88.72\% & \textbf{91.00\%} & 89.77\% \\
XLM-R$_{MLM+TLM}$& 89.19\% & 90.83\% & 89.97\% \\
XLM-R$_{LEM}$ & \textbf{90.63\%} & 90.63\% & \textbf{90.63\%} \\
\hline
\end{tabularx}}
\end{table*}

\section{Ablation Studies}\label{sec:sec_ablation} 
\subsection{The Number of Tokens for Masking in LEM Strategy}\label{subsec:ablation_noof_tokens} 

To evaluate the impact of the masked token count within linguistic entities, we conducted an ablation study by varying the number of masked tokens. We conducted this for the Sinhala-Tamil language pair. As reported in Table~\ref{tab:sita_ablation}, the best result was returned for the 100\% NE+15\% MLM and 100\% NE+15\% TLM combinations in the $LEM_{mono}$ and $LEM_{para}$ steps respectively. Therefore, we used this setting and the number of tokens for masking was varied. Results on the bitext mining task are reported in Table~\ref{tab:lem_token_ablation}. It reveals a clear trend of decreasing performance.

When masked only one token per linguistic entity, the average performance across tasks was the highest. This outcome suggested that minimal masking preserved more contextual information, allowing the model to better capture dependencies critical for downstream tasks. As the masked token count increased to two or more, the average performance dropped. This drop was significant when increasing the token count to 3 and 4. This indicated that excessive masking had disrupted the contextual integrity of linguistic entities, which lead to suboptimal representations.

Interestingly, the performance drop became less pronounced when the number of masked tokens increased from 3 tokens to 4 tokens (a decrease of only 0.02). This suggested a potential saturation point where further masking within an entity had diminishing negative effects, as the model might already struggle to leverage the remaining context effectively.

\begin{table*}[!h]\centering
\caption{Ablation study results by changing the number of tokens masked in the linguistic entity. The results are for the Sinhala-Tamil language pair and the bitext mining downstream task.}
\label{tab:lem_token_ablation}
\scriptsize
\renewcommand{\arraystretch}{1.35}
\resizebox{\textwidth}{!}{%
\begin{tabular}{lcccc}
\toprule
\textbf{No. of Tokens Masked in Linguistic Entity} & \textbf{FW (Recall)} & \textbf{BW (Recall)} & \textbf{IN (Recall)} & \textbf{Average (Recall)} \\
\midrule
1 & \textbf{92.13} & \textbf{93.18} & \textbf{89.10} & \textbf{91.47} \\
2 & 91.02 & 92.52 & 87.78 & 90.44 \\
3 & 83.79 & 87.37 & 79.05 & 83.40 \\
4 & 84.12 & 87.03 & 78.97 & 83.38 \\
\bottomrule
\end{tabular}}
\end{table*}

\subsection{Effect of noise in LEM Strategy}\label{subsec:ablation_noise_indata} 

We investigated the impact of applying the LEM strategy to noisy data. This analysis provides critical insights into the robustness and adaptability of LEM when faced with real-world noisy data. We specifically focus on the English-Sinhala language pair and use the ParaCrawl\footnote{https://opus.nlpl.eu/} dataset.

As per Table~\ref{tab:sien_ablation}, the current ablation revealed that the best results for English-Sinhala were achieved with the combination of 100\%VB+15\%MLM and 100\%VB+15\%TLM during the $LEM_{mono}$ and $LEM_{para}$ steps, respectively. We ran the LEM experiments with ParaCrawl data for the same combinations.

Table~\ref{tab:abl_lem_with_noise} presents the final scores. We observed that the results were comparable to those derived from the cleaner SiTa-Trilingual dataset. Further, in BW criteria the scores slightly surpass and in FW criteria the scores are the same as that obtained when using high-quality data. This equivalence underscores the resilience of the LEM strategy to noise in the training data.

\begin{table*}[!htp]\centering
\caption{Bitext mining results obtained using LEM-enhanced models on both high-quality and noisy web-crawled datasets.}\label{tab:abl_lem_with_noise}
\scriptsize
\renewcommand{\arraystretch}{1.5}
\resizebox{\linewidth}{!}{%
\begin{tabular}{lcrrrrr}\toprule
Dataset &Quality of the Dataset &FW (Recall) &BW (Recall) &IN (Recall) & Overall Average (Recall) \\
\hline
SiTa-Trilingual &High Quality &\textbf{95.09} &94.67 &\textbf{93.42} &\textbf{94.39} \\
ParaCrawl &Noisy &\textbf{95.09} &\textbf{95.00} &92.49 &94.19 \\
\hline
\end{tabular}}
\end{table*}

The ability of LEM to maintain high performance in noisy settings highlights its practical applicability in low-resource scenarios, where parallel data is often noisy or inconsistent. This robustness not only complements our findings but also demonstrates that LEM can effectively mitigate the challenges associated with data quality, a common issue in low-resource language processing.

\section{Discussion}\label{sec:discussion}

The LEM strategy is very much driven by the accuracy of the underlying tools to identify the linguistic entities. The sub-optimal performance of the NER model and POS Taggers can affect the final results.

Although the NER model performs well with English sentences, we observe two main error types with it for Sinhala and Tamil language text, as shown in Table~\ref{tab:ner_limitations} in Appendix~\ref{secD}. As \textit{False Positives}, we observe words which were not a part of the NE tagged as NEs.  Secondly, in the category of \textit{False Negatives}, the NER model fails to identify all the words belonging to the NE sequence, and incorrectly label these words as \textit{Other} etc. Similar instances are found in PoS Tagging as well as per examples in Table~\ref{tab:pos_limitations} in Appendix~\ref{secD} that resulted in \textit{False Positives} and \textit{False Negatives}. However for English language the returned PoS tags were mostly accurate.

As future work, we will continually pre-train a single multilingual model for multiple languages using the LEM strategy. This is in contrast to the current approach which yields specialized encoders for each language pair. Additionally, as a single multilingual model, its performance on downstream tasks can be analysed. Secondly, with the advancements in the field, we hope more sophisticated NER models and PoS Tagger tools might be introduced in future. We will re-evaluate the LEM performance upon the availability of such tools. Thirdly, we will investigate the impact of the LEM strategy on language families, examining its effectiveness across a broader linguistic spectrum.

\clearpage

\section{Conclusion}\label{sec:conclusion} 

Multilingual PLMs trained with masking strategies are less effective for downstream tasks. This research introduced LEM strategy to improve cross-lingual representations of existing multiPLMs. Here the objective is to mask a single token, specifically targeting linguistic entities (NEs, nouns and verbs). Extending this LEM strategy with parallel data yields even better results, as evidenced in low-resource language pairs such as English-Sinhala, English-Tamil, and Sinhala-Tamil. The improved cross-lingual representations showed superior performance on the three evaluation tasks.\\

\noindent\textbf{Funding} Provided upon paper acceptance.\\

\noindent\textbf{Availability of data and materials} We have only used publicly available data in this research. Where applicable, the citations and/or URLs were provided under the relevant sections. \\

\noindent\textbf{Source Code} Will be released upon acceptance.

\begin{appendices}
\section{Type of monolingual data parallel data for MLM}\label{secA}

The Table~\ref{tab:mono_vs_parallel_mlm} shows the results corresponding to Figure~\ref{fig:mono_vs_parallel_mlm}.

\begin{table*}[h]\centering
\caption{Bitext mining recall scores for using pure monolingual data versus source and target sides from a parallel corpus (as monolingual data) for MLM experiments.}\label{tab:mono_vs_parallel_mlm}
\scriptsize
\renewcommand{\arraystretch}{1.5}
\resizebox{\linewidth}{!}{%
\begin{tabular}{lrrrrrrrrrrrrrrrrr}\toprule
\multirow{2}{*}{\textbf{Dataset}} &\multirow{2}{*}{\textbf{Dataset Size}} &\multicolumn{3}{c}{\textbf{Army}} &\multicolumn{3}{c}{\textbf{Hiru}} &\multicolumn{3}{c}{\textbf{ITN}} &\multicolumn{3}{c}{\textbf{Newsfirst}} &\multicolumn{3}{c}{\textbf{Averages}} \\\cmidrule{3-17}
& &\textbf{F} &\textbf{B} &\textbf{I} &\textbf{F} &\textbf{B} &\textbf{I} &\textbf{F} &\textbf{B} &\textbf{I} &\textbf{F} &\textbf{B} &\textbf{I} &\textbf{F} &\textbf{B} &\textbf{I} \\
\midrule
\multicolumn{17}{c}{\textbf{English - Sinhala }} \\
\midrule
SiTa &59333 &88.33 &91.00 &85.33 &92.03 &93.36 &89.70 &91.67 &92.67 &88.67 &91.67 &95.33 &90.00 &\textbf{90.92} &\textbf{93.09} &\textbf{88.42} \\
MADLAD400 &60000 &82.67 &88.33 &78.00 &85.05 &91.36 &82.00 &85.33 &86.00 &79.67 &91.67 &92.67 &87.67 &86.18 &89.59 &81.83 \\
MADLAD400 &100000 &86.67 &91.67 &83.33 &91.69 &96.01 &91.03 &88.00 &90.33 &83.00 &91.33 &95.00 &89.00 &89.42 &93.25 &86.59 \\
\midrule
\multicolumn{17}{c}{\textbf{English - Tamil }} \\
\midrule
SiTa &59333 &84.00 &86.00 &77.67 &80.33 &75.00 &68.33 &81.56 &82.21 &78.52 &90.67 &91.00 &87.00 &\textbf{84.14} &\textbf{83.55} &\textbf{77.88} \\
MADLAD400 &60000 &81.67 &78.67 &69.33 &75.33 &69.67 &60.67 &81.18 &77.15 &69.77 &90.00 &86.67 &81.67 &82.05 &78.04 &70.36 \\
MADLAD400 &100000 &81.33 &79.67 &71.67 &77.67 &71.33 &62.67 &78.86 &76.17 &68.79 &88.67 &88.00 &82.00 &81.63 &78.79 &71.28 \\
\midrule
\multicolumn{17}{c}{\textbf{Sinhala - Tamil}} \\
\midrule
SiTa &59333 &86.75 &88.08 &81.46 &88.00 &89.33 &84.00 &93.33 &92.67 &89.33 &90.33 &94.00 &89.00 &\textbf{89.60} &\textbf{91.02} &\textbf{85.95} \\
MADLAD400 &60000 &84.77 &89.73 &80.46 &86.00 &89.00 &83.00 &92.67 &92.00 &89.00 &89.00 &92.67 &85.67 &88.11 &90.85 &84.53 \\
MADLAD400 &100000 &84.11 &88.08 &78.81 &86.00 &89.33 &81.33 &90.67 &93.67 &87.33 &88.67 &92.33 &85.00 &87.36 &90.85 &83.12 \\
MADLAD400 &500000 &82.12 &83.11 &75.17 &85.67 &88.33 &79.67 &87.67 &91.00 &83.67 &87.67 &90.67 &82.33 &85.78 &88.28 &80.21 \\
\bottomrule
\end{tabular}
}
\end{table*}

\section{Parallel Data Curation Task NMT extrinsic Evaluation Results}\label{secB}

The NMT evaluation scores for the parallel data curation task are reported in Table~\ref{tab:pdc_nmt_scores} for the Flores+  benchmark devtest evaluation set. While the discussion on the NMT results has been based using the ChrF++ metric, here we present the scores for the same experiments using NMT evaluation metrics sacreBLEU, multi-bleu, ChrF, ChrF++ and spBLEU. 

\begin{table*}[h]\centering
\caption{NMT scores on the Flores+ devtest using top 50,000 parallel sentences from the ranked NLLB and CCAligned corpus.}\label{tab:pdc_nmt_scores}
\scriptsize
\renewcommand{\arraystretch}{1.35}
\resizebox{0.8\linewidth}{!}{%
\begin{tabular}{lrrrrrr}\toprule
\textbf{} &\textbf{sacreBLEU} &\textbf{multi-bleu} &\textbf{ChrF} &\textbf{ChrF++} &\textbf{SpBLEU} \\
\hline
\multicolumn{6}{c}{\textbf{NLLB}} \\\midrule
\multicolumn{6}{c}{\textbf{Sinhala - Tamil}} \\
XLM-R &2.6 &2.58 &38.6 &33.58 &11.9 \\
XLM-R$_{MLM+TLM}$ &3.2 &3.23 &41.3 &35.99 &14.5 \\
XLM-R$_{LEM}$ &\textbf{3.6} &\textbf{3.60} &\textbf{42.1} &\textbf{36.68} &\textbf{15.2} \\
\multicolumn{6}{c}{\textbf{English - Tamil}} \\
XLM-R &6.2 &6.18 &44.00 &38.28 &18.40 \\
XLM-R$_{MLM+TLM}$ &9.2 &9.16 &50.70 &45.35 &25.20 \\
XLM-R$_{LEM}$ &\textbf{9.3} &\textbf{9.47} &\textbf{51.20} &\textbf{45.86} &\textbf{25.80} \\
\multicolumn{6}{c}{\textbf{English - Sinhala}} \\
XLM-R &4.9 &4.91 &33.1 &30.37 &13.6 \\
XLM-R$_{MLM+TLM}$ &9.4 &9.42 &43.2 &39.78 &23.3 \\
XLM-R$_{LEM}$ &\textbf{9.9} &\textbf{9.85} &\textbf{43.9} &\textbf{40.31} &\textbf{23.8}\\
\hline
\multicolumn{6}{c}{\textbf{CCAligned}}\\
\hline
\multicolumn{6}{c}{\textbf{Sinhala - Tamil}} \\
XLM-R &2.2 &2.23 &37.2 &32.43 &10.6 \\
XLM-R$_{MLM+TLM}$ &3.7 &3.74 &42.3 &36.02 &15.2 \\
XLM-R$_{LEM}$ &\textbf{3.6} &\textbf{3.61} &\textbf{42.6} &\textbf{36.90} &\textbf{14.9}\\
\multicolumn{6}{c}{\textbf{English - Tamil}} \\
XLM-R &0.2 &0.17 &5.2 &5.80 &1.2\\
XLM-R$_{MLM+TLM}$ &3.2 &3.24 &31.5 &28.55 &11.5\\
XLM-R$_{LEM}$ &\textbf{3.5} &\textbf{3.48} &\textbf{34.3} &\textbf{30.96} &\textbf{12.5} \\
\multicolumn{6}{c}{\textbf{English - Sinhala}} \\
XLM-R &0.4 &0.37 &10.2 &10.13 &2.3 \\
XLM-R$_{MLM+TLM}$ &5.0 &5.00 &33.9 &31.17 &14.8 \\
XLM-R$_{LEM}$ &\textbf{5.1} &\textbf{5.09} &\textbf{34.5} &\textbf{31.71} &\textbf{15.3} \\
\bottomrule
\end{tabular}}
\end{table*}

\section{Linguistic Entity Masking Ablation Study}\label{secC}

In this section, we present the full experiments along with the scores obtained during the ablation study of our LEM masking strategy. The Tables~\ref{tab:sien_ablation}, \ref{tab:taen_ablation} and \ref{tab:sita_ablation} contain the bitext mining recall scores for English-Sinhala, English-Tamil and Sinhala-Tamil language pairs respectively.

\begin{sidewaystable*}[p]\centering
\caption{Ablation experiments and bitext mining scores for English-Sinhala language pair considering linguistic entity masking.}\label{tab:sien_ablation}
\scriptsize
\resizebox{\linewidth}{!}{%
\begin{tabular}{lrrrrrrrrrrrrrrrr}\toprule
\textbf{Experiment} &\multicolumn{3}{c}{\textbf{Army}} &\multicolumn{3}{c}{\textbf{Hiru}} &\multicolumn{3}{c}{\textbf{ITN}} &\multicolumn{3}{c}{\textbf{Newsfirst}} &\multicolumn{3}{c}{\textbf{Averages}} \\\cmidrule{1-16}
\textbf{} &\textbf{F} &\textbf{B} &\textbf{I} &\textbf{F} &\textbf{B} &\textbf{I} &\textbf{F} &\textbf{B} &\textbf{I} &\textbf{F} &\textbf{B} &\textbf{I} &\textbf{F} &\textbf{B} &\textbf{I} \\
\midrule
\multicolumn{16}{l}{\textbf{Baselines}} \\
XLM-R &92.33 &93.33 &89.67 &96.35 &96.68 &\colorbox{yellow}{\textbf{95.68}} &94.00 &96.00 &92.33 &96.67 &95.33 &\colorbox{yellow}{\textbf{94.33}} &94.84 &95.34 &93.00 \\
15\%MLM &88.33 &91.00 &85.33 &92.03 &93.36 &89.70 &91.67 &92.67 &88.67 &91.67 &95.33 &90.00 &90.92 &93.09 &88.42 \\
15\% TLM on 15\% MLM &91.33 &92.67 &88.67 &94.35 &95.68 &93.36 &94.00 &94.00 &90.67 &94.67 &95.00 &92.67 &93.59 &94.34 &91.34 \\
\hline
\multicolumn{16}{l}{\textbf{LEM$_{mono}$}} \\
100\%NE+15\% MLM &89.67 &93.00 &88.33 &93.02 &94.02 &92.03 &89.67 &93.00 &87.00 &93.67 &94.67 &91.67 &91.51 &93.67 &89.76 \\
100\% VB+15\% MLM &89.67 &93.33 &87.33 &94.02 &95.02 &92.69 &92.00 &93.67 &89.67 &93.00 &95.33 &92.33 &92.17 &94.34 &90.51 \\
100\% NN+15\% MLM &81.33 &88.33 &76.33 &93.36 &95.02 &92.36 &90.33 &91.67 &86.00 &91.00 &92.33 &87.67 &89.00 &91.84 &85.59 \\
100\% NE+ 100\%VB+15\% MLM &91.33 &91.00 &87.67 &95.35 &94.02 &93.36 &92.33 &94.00 &89.33 &93.33 &94.33 &90.67 &93.09 &93.34 &90.26 \\
100\% NE+ 100\%NN+15\% MLM &88.00 &91.00 &84.00 &94.02 &95.35 &92.69 &89.33 &95.67 &89.00 &94.00 &95.67 &91.67 &91.34 &94.42 &89.34 \\
100\% NE+ 100\%VB+ 100\%NN+15\% MLM &89.67 &92.33 &87.00 &94.02 &94.02 &91.69 &92.33 &95.00 &91.00 &94.00 &92.33 &90.33 &92.50 &93.42 &90.01 \\
\hline
\multicolumn{16}{l}{\textbf{MLM$_{mono}$+TLM$_{para}$}} \\
100\% NE+15\% TLM on 15\% MLM &90.00 &91.67 &87.33 &95.02 &95.35 &93.36 &94.00 &\colorbox{yellow}{\textbf{96.67}} &92.67 &96.67 &96.67 &93.33 &93.92 &95.09 &91.67 \\
100\% VB+15\% TLM on 15\% MLM &91.67 &90.33 &86.67 &94.35 &95.02 &92.69 &93.00 &95.33 &89.67 &95.00 &94.67 &91.67 &93.50 &93.84 &90.17 \\
100\% NN+15\% TLM on 15\% MLM &89.00 &92.00 &85.00 &93.36 &95.02 &91.36 &94.33 &96.00 &92.33 &94.67 &95.00 &92.00 &92.84 &94.50 &90.17 \\
100\% NE+ 100\%VB+15\% TLM on 15\% MLM &91.33 &91.33 &87.67 &95.35 &94.68 &92.69 &94.00 &95.00 &91.33 &\colorbox{yellow}{\textbf{97.33}} &95.00 &93.67 &94.50 &94.00 &91.34 \\
100\% NE+ 100\%NN+15\% TLM on 15\% MLM &88.67 &91.00 &85.00 &94.35 &95.35 &93.02 &94.00 &96.00 &92.00 &93.67 &95.00 &91.33 &92.67 &94.34 &90.34 \\
100\%NE+100\%VB+100\%NN+15\%TLM on 15\% MLM &90.67 &91.33 &87.33 &94.68 &\colorbox{yellow}{\textbf{97.34}} &94.35 &93.67 &95.00 &91.00 &94.33 &96.33 &92.33 &93.34 &95.00 &91.25 \\
\multicolumn{16}{c}{\textbf{}} \\
15\% TLM on (100\%NE+15\% MLM) &89.00 &93.00 &87.00 &94.35 &95.35 &93.64 &92.00 &95.67 &90.00 &95.00 &90.00 &93.33 &92.59 &93.50 &90.99 \\
100\% NE+15\% TLM on (100\%NE+15\% MLM) &91.67 &\colorbox{yellow}{\textbf{95.33}} &89.33 &94.68 &96.01 &94.35 &92.00 &96.33 &92.67 &94.67 &95.67 &92.67 &93.25 &\colorbox{yellow}{\textbf{95.84}} &92.25 \\
100\% VB+15\% TLM on (100\%NE+15\% MLM) &90.00 &91.67 &86.00 &94.02 &95.02 &93.36 &92.67 &94.67 &90.00 &93.33 &95.00 &91.67 &92.50 &94.09 &90.26 \\
100\% NN+15\% TLM on (100\%NE+15\% MLM) &89.00 &92.00 &87.00 &94.02 &94.02 &92.36 &93.00 &93.33 &89.00 &94.00 &94.00 &91.00 &92.50 &93.34 &89.84 \\
100\% NE+ 100\%VB+15\% TLM on (100\%NE+15\% MLM) &89.67 &93.33 &88.00 &95.02 &94.68 &93.36 &92.00 &95.33 &90.00 &95.67 &95.33 &93.33 &93.09 &94.67 &91.17 \\
100\% NE+ 100\%NN+15\% TLM on (100\%NE+15\% MLM) &89.33 &93.00 &87.00 &94.35 &94.68 &93.02 &93.67 &94.67 &90.67 &95.67 &96.67 &94.00 &93.25 &94.75 &91.17 \\
100\%NE+100\%VB+100\%NN+15\%TLM on (100\%NE+15\% MLM) &91.67 &92.33 &88.33 &95.68 &95.68 &95.02 &92.33 &93.33 &88.67 &93.67 &95.00 &91.33 &93.34 &94.09 &90.84 \\
\multicolumn{16}{c}{\textbf{}} \\
15\% TLM on (100\%VB+15\% MLM) &91.67 &92.00 &89.00 &94.35 &96.01 &94.02 &94.33 &95.00 &91.67 &95.67 &96.00 &93.33 &94.00 &94.75 &92.00 \\
100\% NE+15\% TLM on (100\%VB+15\% MLM) &90.33 &91.67 &87.67 &95.02 &96.35 &94.35 &93.67 &94.33 &90.00 &96.67 &95.67 &93.67 &93.92 &94.50 &91.42 \\
100\% VB+15\% TLM on (100\%VB+15\% MLM) &91.67 &93.33 &90.67 &\colorbox{yellow}{\textbf{96.68}} &95.35 &95.35 &95.33 &94.33 &93.67 &96.67 &95.67 &94.00 &\colorbox{yellow}{\textbf{95.09}} &94.67 &\colorbox{yellow}{\textbf{93.42}} \\
100\% NN+15\% TLM on (100\%VB+15\% MLM) &88.33 &91.67 &86.00 &95.02 &95.02 &93.36 &93.00 &93.67 &89.67 &92.67 &94.67 &91.33 &92.25 &93.75 &90.09 \\
100\% NE+ 100\%VB+15\% TLM on (100\%VB+15\% MLM) &90.00 &94.33 &88.33 &94.35 &95.68 &93.02 &94.00 &95.33 &91.00 &96.67 &95.33 &\colorbox{yellow}{\textbf{94.33}} &93.75 &95.17 &91.67 \\
100\% NE+ 100\%NN+15\% TLM on (100\%VB+15\% MLM) &89.67 &91.33 &86.33 &94.68 &95.68 &93.69 &93.67 &94.33 &91.33 &95.67 &96.33 &93.67 &93.42 &94.42 &91.26 \\
100\%NE+100\%VB+100\%NN+15\%TLM on (100\%VB+15\% MLM) &92.00 &92.33 &87.33 &95.35 &95.68 &94.02 &93.00 &94.00 &89.67 &95.67 &95.00 &93.00 &94.00 &94.25 &91.00 \\
\multicolumn{16}{c}{\textbf{}} \\
15\% TLM on (100\%NN+15\%MLM) &90.33 &93.33 &87.33 &94.35 &94.68 &93.02 &94.67 &95.00 &92.00 &94.67 &95.00 &92.67 &93.50 &94.50 &91.26 \\
100\% NE+15\% TLM on (100\%NN+15\%MLM) &89.00 &93.67 &87.00 &94.35 &95.35 &92.36 &95.00 &95.33 &91.33 &96.00 &95.33 &92.67 &93.59 &94.92 &90.84 \\
100\% VB+15\% TLM on (100\%NN+15\%MLM) &88.00 &93.33 &86.67 &93.69 &95.68 &93.02 &94.33 &95.67 &\colorbox{yellow}{\textbf{94.67}} &94.67 &94.00 &91.67 &92.67 &94.67 &91.51 \\
100\% NN+15\% TLM on (100\%NN+15\%MLM) &91.00 &92.00 &87.67 &95.68 &95.02 &94.02 &94.33 &96.33 &92.67 &95.00 &95.67 &92.67 &94.00 &94.75 &91.76 \\
100\% NE+ 100\%VB+15\% TLM on (100\%NN+15\%MLM) &90.67 &93.67 &87.67 &95.02 &94.68 &93.02 &95.00 &95.67 &92.33 &94.33 &94.00 &91.67 &93.75 &94.50 &91.17 \\
100\% NE+ 100\%NN+15\% TLM on (100\%NN+15\%MLM) &91.67 &91.33 &87.67 &94.68 &95.68 &94.02 &93.00 &95.33 &90.67 &94.33 &95.00 &92.33 &93.42 &94.34 &91.17 \\
100\%NE+100\%VB+100\%NN+15\%TLM on (100\%NN+15\%MLM) &88.67 &92.00 &86.00 &96.01 &96.01 &95.02 &94.00 &95.33 &91.33 &94.33 &94.67 &91.33 &93.25 &94.50 &90.92 \\
\multicolumn{16}{c}{\textbf{}} \\
15\% TLM on (100\%NE+100\%VB+15\%MLM) &88.67 &93.00 &86.67 &94.35 &95.02 &93.02 &92.33 &93.67 &88.67 &93.00 &94.33 &90.67 &92.09 &94.00 &89.76 \\
100\% NE+15\% TLM on (100\%NE+100\%VB+15\%MLM) &89.67 &91.33 &87.00 &94.68 &95.68 &93.69 &93.67 &94.33 &93.67 &95.67 &94.33 &91.33 &93.42 &93.92 &91.42 \\
100\% VB+15\% TLM on (100\%NE+100\%VB+15\%MLM) &88.00 &93.33 &86.67 &93.69 &95.68 &93.02 &94.33 &94.33 &\colorbox{yellow}{\textbf{94.67}} &94.67 &94.00 &91.67 &92.67 &94.34 &91.51 \\
100\% NN+15\% TLM on (100\%NE+100\%VB+15\%MLM) &88.67 &93.00 &86.67 &93.67 &95.02 &93.02 &94.00 &93.67 &90.00 &93.00 &94.33 &90.67 &92.33 &94.00 &90.09 \\
100\% NE+ 100\%VB+15\% TLM on (100\%NE+100\%VB+15\%MLM) &91.33 &92.67 &89.00 &94.68 &95.68 &93.69 &94.00 &94.67 &91.67 &95.33 &95.33 &93.00 &93.84 &94.59 &91.84 \\
100\% NE+ 100\%NN+15\% TLM on (100\%NE+100\%VB+15\%MLM) &91.00 &91.33 &87.67 &94.02 &94.68 &92.36 &94.67 &94.67 &91.67 &95.67 &95.33 &93.00 &93.84 &94.00 &91.17 \\
100\%NE+100\%VB+100\%NN+15\%TLM on (100\%NE+100\%VB+15\%MLM) &92.00 &93.00 &89.00 &95.35 &96.01 &94.68 &94.33 &94.00 &91.00 &96.00 &96.00 &\colorbox{yellow}{\textbf{94.33}} &94.42 &94.75 &92.25 \\
\multicolumn{16}{c}{\textbf{}} \\
15\% TLM on (100\%NE+100\%NN+15\%MLM) &91.33 &94.00 &88.67 &94.02 &95.02 &92.03 &95.33 &95.67 &93.00 &94.33 &\colorbox{yellow}{\textbf{97.67}} &94.00 &93.75 &95.59 &91.92 \\
100\% NE+15\% TLM on (100\%NE+100\%NN+15\%MLM) &87.67 &90.33 &\colorbox{yellow}{\textbf{93.67}} &94.02 &95.35 &93.02 &\colorbox{yellow}{\textbf{96.00}} &94.67 &92.67 &93.67 &94.67 &91.67 &92.84 &93.75 &92.76 \\
100\% VB+15\% TLM on (100\%NE+100\%NN+15\%MLM) &91.00 &92.00 &87.00 &94.35 &94.35 &92.69 &93.67 &96.33 &93.00 &95.67 &95.33 &93.00 &93.67 &94.50 &91.42 \\
100\% NN+15\% TLM on (100\%NE+100\%NN+15\%MLM) &88.33 &91.67 &84.33 &95.02 &95.35 &93.64 &94.67 &94.67 &91.67 &94.33 &95.33 &92.33 &93.09 &94.25 &90.49 \\
100\% NE+100\%VB+15\% TLM on (100\%NE+100\%NN+15\%MLM) &90.00 &93.33 &86.67 &94.68 &94.68 &93.36 &93.33 &93.00 &89.33 &96.33 &96.00 &\colorbox{yellow}{\textbf{94.33}} &93.59 &94.25 &90.92 \\
100\% NE+100\%NN+15\% TLM on (100\%NE+100\%NN+15\%MLM) &87.67 &90.33 &84.00 &95.68 &96.01 &94.35 &92.00 &95.00 &90.33 &94.67 &96.33 &93.00 &92.50 &94.42 &90.42 \\
100\%NE+100\%VB+100\%NN+15\%TLM on (100\%NE+100\%NN+15\%MLM) &88.67 &91.67 &85.00 &95.35 &95.68 &94.35 &94.00 &93.67 &90.33 &95.67 &95.33 &93.00 &93.42 &94.09 &90.67 \\
\multicolumn{16}{c}{\textbf{}} \\
15\% TLM on (100\%NE+100\%VB+100\%NN+15\%MLM) &\colorbox{yellow}{\textbf{93.00}} &91.67 &88.00 &95.35 &96.01 &94.02 &94.67 &95.00 &92.00 &93.67 &94.67 &91.67 &94.17 &94.34 &91.42 \\
100\% NE+15\% TLM on (100\%NE+100\%VB+100\%NN+15\%MLM) &89.00 &91.00 &84.67 &95.35 &96.35 &94.68 &\colorbox{yellow}{\textbf{96.00}} &95.33 &93.00 &96.00 &95.67 &93.33 &94.09 &94.59 &91.42 \\
100\% VB+15\% TLM on (100\%NE+100\%VB+100\%NN+15\%MLM) &89.67 &92.67 &87.00 &95.35 &95.35 &93.67 &95.00 &93.33 &91.67 &95.67 &93.33 &91.00 &93.92 &93.67 &90.83 \\
100\% NN+15\% TLM on (100\%NE+100\%VB+100\%NN+15\%MLM) &89.67 &91.33 &86.00 &95.02 &95.35 &93.33 &93.67 &94.67 &91.33 &93.67 &94.00 &90.33 &93.00 &93.84 &90.25 \\
100\% NE+ 100\%VB+15\% TLM on (100\%NE+100\%VB+100\%NN+15\%MLM) &88.67 &92.00 &85.33 &93.69 &95.68 &92.69 &93.00 &95.67 &90.33 &95.00 &95.33 &92.33 &92.59 &94.67 &90.17 \\
100\% NE+ 100\%NN+15\% TLM on (100\%NE+100\%VB+100\%NN+15\%MLM) &86.67 &91.67 &84.67 &96.35 &96.01 &94.68 &94.33 &95.67 &92.33 &94.33 &94.67 &91.33 &92.92 &94.50 &90.75 \\
100\%NE+100\%VB+100\%NN+15\%TLM on (100\%NE+100\%VB+100\%NN+15\%MLM) &91.33 &92.00 &88.00 &95.68 &95.68 &94.02 &93.67 &94.00 &91.33 &93.67 &94.00 &90.33 &93.59 &93.92 &90.92 \\
\bottomrule
\end{tabular}
}
\end{sidewaystable*}

\begin{sidewaystable*}[p]\centering
\caption{Ablation experiments and bitext mining scores for English-Tamil language pair considering linguistic entity masking.}\label{tab:taen_ablation}
\scriptsize
\resizebox{\linewidth}{!}{%
\begin{tabular}{lrrrrrrrrrrrrrrrr}\toprule
\multirow{2}{*}{\textbf{Experiment}} &\multicolumn{3}{c}{\textbf{Army}} &\multicolumn{3}{c}{\textbf{Hiru}} &\multicolumn{3}{c}{\textbf{ITN}} &\multicolumn{3}{c}{\textbf{Newsfirst}} &\multicolumn{3}{c}{\textbf{Average}} \\\cmidrule{2-16}
&\textbf{FW} &\textbf{BW} &\textbf{IN} &\textbf{FW} &\textbf{BW} &\textbf{IN} &\textbf{FW} &\textbf{BW} &\textbf{IN} &\textbf{FW} &\textbf{BW} &\textbf{IN} &\textbf{FW} &\textbf{BW} &\textbf{IN} \\\midrule
\multicolumn{16}{l}{\textbf{Baselines}} \\
XLM-R &86.67 &88.33 &82.00 &83.00 &78.33 &72.67 &83.22 &83.56 &78.86 &92.33 &91.33 &89.33 &86.31 &85.39 &80.71 \\
15\%MLM &84.00 &86.00 &77.67 &80.33 &75.00 &68.33 &81.56 &82.21 &78.52 &90.67 &91.00 &87.00 &84.14 &83.55 &77.88 \\
15\%TLM on 15\%MLM &86.67 &85.67 &79.33 &80.33 &78.67 &71.00 &81.88 &83.56 &77.52 &90.00 &92.67 &88.00 &84.72 &85.14 &78.96 \\
\hline
\multicolumn{16}{l}{\textbf{LEM$_{mono}$}} \\
100\% NE+15\% MLM &86.00 &86.67 &81.00 &79.33 &75.33 &66.67 &81.21 &81.21 &74.83 &93.00 &92.00 &90.00 &84.89 &83.80 &78.12 \\
100\% VB+15\% MLM &85.67 &84.67 &76.67 &78.67 &76.00 &68.00 &81.88 &82.55 &75.84 &91.00 &90.00 &86.33 &84.30 &83.30 &76.71 \\
100\% NN+15\% MLM &83.33 &84.67 &77.00 &73.67 &72.67 &61.67 &75.84 &82.22 &70.13 &90.00 &91.00 &87.00 &80.71 &82.64 &73.95 \\
100\% NE+100\%VB+15\% MLM &83.00 &86.67 &77.67 &77.67 &74.33 &65.00 &81.21 &83.56 &75.84 &89.00 &88.67 &84.00 &82.72 &83.31 &75.63 \\
100\% NE+100\%NN+15\% MLM &82.67 &85.33 &75.00 &75.33 &72.67 &62.00 &80.54 &84.23 &74.48 &90.33 &90.00 &86.33 &82.22 &83.06 &74.45 \\
100\% NE+100\% VB +100\%NN+15\% MLM &83.00 &83.33 &78.00 &74.67 &73.67 &64.67 &80.87 &83.89 &76.17 &91.33 &92.67 &88.67 &82.47 &83.39 &76.88 \\
\hline
\multicolumn{16}{l}{\textbf{LEM$_{mono}$+LEM$_{para}$}} \\
100\% NE+15\% TLM on 15\%MLM &83.00 &85.33 &76.33 &79.67 &78.33 &70.00 &83.89 &85.91 &79.87 &91.00 &93.33 &89.00 &84.39 &85.73 &78.80 \\
100\% VB+15\% TLM on 15\%MLM &87.00 &86.67 &81.67 &80.67 &79.00 &72.33 &83.89 &85.57 &79.87 &91.67 &92.33 &88.67 &85.81 &85.89 &80.63 \\
100\% NN+15\% TLM on 15\%MLM &85.00 &86.67 &79.67 &79.33 &77.00 &69.00 &83.89 &86.24 &80.54 &91.33 &94.00 &89.67 &84.89 &85.98 &79.72 \\
100\% NE+100\% VB+15\% TLM on 15\%MLM &85.67 &76.67 &69.33 &79.67 &76.67 &69.33 &83.22 &84.23 &77.85 &92.00 &92.00 &89.33 &85.14 &82.39 &76.46 \\
100\% NE+100\% NN+15\% TLM on 15\%MLM &84.67 &85.00 &77.67 &81.00 &80.00 &72.33 &81.98 &84.23 &77.85 &90.00 &92.00 &87.67 &84.41 &85.31 &78.88 \\
100\% NE+100\% VB+ 100\% NN+ 15\% TLM on 15\%MLM &85.00 &85.33 &80.00 &78.67 &78.33 &70.00 &84.23 &\colorbox{yellow}{\textbf{88.59}} &\colorbox{yellow}{\textbf{80.87}} &90.00 &93.67 &88.33 &84.47 &86.48 &79.80 \\
\multicolumn{16}{c}{\textbf{}} \\
15\% TLM on 100\%NE+15\%MLM &87.00 &86.33 &81.33 &81.33 &80.00 &71.67 &81.21 &84.23 &77.52 &92.67 &91.33 &89.00 &85.55 &85.47 &79.88 \\
100\%NE+15\% TLM on 100\%NE+15\%MLM &87.67 &87.00 &81.67 &82.00 &\colorbox{yellow}{\textbf{81.33}} &\colorbox{yellow}{\textbf{73.00}} &81.88 &84.23 &77.18 &91.33 &92.67 &88.33 &85.72 &86.31 &80.05 \\
100\% VB+15\% TLM on 100\%NE+15\%MLM &\colorbox{yellow}{\textbf{88.33}} &89.33 &\colorbox{yellow}{\textbf{83.67}} &80.00 &77.67 &69.67 &81.54 &84.23 &75.50 &91.00 &93.00 &89.00 &85.22 &86.06 &79.46 \\
100\% NN+15\% TLM on 100\%NE+15\%MLM &86.33 &87.67 &80.33 &81.33 &79.67 &70.67 &80.54 &84.56 &76.51 &92.00 &91.33 &88.00 &85.05 &85.81 &78.88 \\
100\% NE+100\% VB+15\% TLM on 100\%NE+15\%MLM/ &84.67 &85.67 &78.00 &82.33 &76.67 &70.33 &80.54 &83.22 &76.85 &89.33 &92.67 &87.67 &84.22 &84.56 &78.21 \\
100\% NE+100\% NN+15\% TLM on 100\%NE+15\%MLM/ &84.67 &85.67 &78.00 &82.33 &76.67 &70.33 &80.54 &83.22 &76.85 &89.33 &92.67 &87.67 &84.22 &84.56 &78.21 \\
100\%NE+100\%VB+100\%NN+15\%TLM on 100\%NE+15\%MLM/ &85.00 &84.33 &78.33 &78.00 &76.67 &67.33 &79.53 &83.89 &75.84 &92.33 &92.00 &90.00 &83.72 &84.22 &77.88 \\
\multicolumn{16}{c}{\textbf{}} \\
15\% TLM on (100\%VB+15\%MLM) &88.00 &88.67 &83.67 &82.00 &79.00 &72.33 &84.90 &84.90 &80.54 &\colorbox{yellow}{\textbf{93.33}} &93.00 &\colorbox{yellow}{\textbf{91.00}} &\colorbox{yellow}{\textbf{87.06}} &86.39 &\colorbox{yellow}{\textbf{81.88}} \\
100\% NE+ 15\% TLM on (100\%VB+15\%MLM) &84.00 &87.67 &79.00 &78.67 &\colorbox{yellow}{\textbf{81.33}} &71.00 &82.22 &85.57 &78.86 &90.67 &93.33 &88.33 &83.89 &\colorbox{yellow}{\textbf{86.98}} &79.30 \\
100\% VB+ 15\% TLM on (100\%VB+15\%MLM) &86.00 &88.67 &80.67 &81.33 &78.33 &70.67 &82.22 &84.56 &76.85 &91.33 &92.33 &87.67 &85.22 &85.97 &78.96 \\
100\% NN+15\% TLM on (100\%VB+15\%MLM) &86.33 &85.33 &80.33 &79.67 &79.00 &70.33 &82.22 &84.90 &77.52 &90.33 &93.67 &88.00 &84.64 &85.72 &79.05 \\
100\% NE+ 100\% VB+ 15\% TLM on (100\%VB+15\%MLM) &85.67 &88.00 &80.33 &80.33 &76.00 &69.00 &81.54 &83.58 &77.18 &90.67 &93.00 &88.00 &84.55 &85.14 &78.63 \\
100\% NE+ 100\% NN+ 15\% TLM on (100\%VB+15\%MLM) &87.33 &87.67 &81.67 &78.00 &78.00 &68.67 &81.54 &83.89 &76.85 &91.00 &92.00 &87.67 &84.47 &85.39 &78.71 \\
100\% NE+ 100\% NN+ 100\%VB+ 15\% TLM on (100\%VB+15\%MLM) &86.33 &87.00 &80.67 &78.33 &76.67 &67.33 &82.89 &84.56 &77.85 &90.67 &92.33 &87.33 &84.55 &85.14 &78.30 \\
\multicolumn{16}{c}{\textbf{}} \\
15\% TLM on (100\%NN+15\%MLM) &84.67 &88.33 &81.00 &81.00 &77.33 &69.67 &83.22 &85.91 &78.86 &91.67 &92.33 &89.67 &85.14 &85.98 &79.80 \\
100\% NE+ 15\% TLM on (100\%NN+15\%MLM) &85.33 &86.67 &79.00 &78.33 &76.33 &67.33 &81.88 &84.56 &76.85 &91.00 &91.33 &87.67 &84.14 &84.72 &77.71 \\
100\% VB+15\% TLM on (100\%NN+15\%MLM) &84.67 &87.67 &80.67 &78.67 &76.00 &67.33 &79.19 &84.29 &75.50 &90.00 &92.67 &88.00 &83.13 &85.16 &77.88 \\
100\% NN+ 15\% TLM on (100\%NN+15\%MLM) &85.00 &87.00 &79.33 &77.33 &76.33 &66.00 &80.87 &83.56 &74.83 &89.67 &92.67 &87.00 &83.22 &84.89 &76.79 \\
100\% NE+ 100\% VB+ 15\% TLM on (100\%NN+15\%MLM) &82.33 &86.00 &77.67 &78.33 &74.33 &65.00 &81.21 &85.91 &76.51 &62.67 &65.00 &61.00 &76.14 &77.81 &70.04 \\
100\% NE+ 100\% NN+ 15\% TLM on (100\%NN+15\%MLM) &86.00 &87.00 &80.00 &78.00 &76.67 &66.67 &79.53 &84.23 &74.16 &88.67 &92.33 &86.67 &83.05 &85.06 &76.87 \\
100\% NE+ 100\% NN+ 100\%VB+ 15\% TLM on (100\%NN+15\%MLM) &86.33 &\colorbox{yellow}{\textbf{90.00}} &82.00 &76.00 &78.33 &66.33 &79.87 &85.91 &76.16 &90.33 &92.00 &87.67 &83.13 &86.56 &78.04 \\
\multicolumn{16}{c}{\textbf{}} \\
15\% TLM on (100\%NE+100\%VB+15\%MLM) &85.33 &89.33 &80.00 &80.00 &75.33 &67.33 &85.34 &84.29 &78.86 &90.00 &91.67 &87.00 &85.17 &85.16 &78.30 \\
100\% NE+ 15\% TLM on (100\%NE+100\%VB+15\%MLM) &86.00 &87.33 &80.33 &80.33 &78.33 &71.33 &82.22 &81.88 &75.50 &89.00 &93.00 &88.00 &84.39 &85.14 &78.79 \\
100\% VB+15\% TLM on (100\%NE+100\%VB+15\%MLM) &84.67 &87.67 &79.33 &78.00 &75.33 &67.00 &83.89 &84.56 &77.85 &88.33 &92.00 &86.67 &83.72 &84.89 &77.71 \\
100\% NN+ 15\% TLM on (100\%NE+100\%VB+15\%MLM) &86.00 &86.67 &79.00 &81.00 &75.67 &67.00 &83.89 &84.56 &78.52 &92.00 &93.00 &89.33 &85.72 &84.97 &78.46 \\
100\% NE+100\% VB+ 15\% TLM on (100\%NE+100\%VB+15\%MLM) &84.67 &87.67 &78.00 &78.33 &75.67 &68.00 &\colorbox{yellow}{\textbf{90.87}} &84.90 &76.17 &89.00 &92.00 &86.67 &85.72 &85.06 &77.21 \\
100\% NE+ 100\% NN+ 15\% TLM on (100\%NE+100\%VB+15\%MLM) &83.00 &86.67 &78.33 &\colorbox{yellow}{\textbf{84.67}} &77.00 &72.00 &81.88 &84.56 &77.18 &89.00 &93.00 &87.67 &84.64 &85.31 &78.80 \\
100\%NE+100\%VB+100\%NN+15\%TLM on (100\%NE+100\%VB+15\%MLM) &87.67 &86.33 &80.00 &79.00 &78.00 &69.33 &82.89 &84.29 &78.52 &90.00 &93.67 &88.67 &84.89 &85.57 &79.13 \\
\multicolumn{16}{c}{\textbf{}} \\
15\% TLM on (100\%NE+100\%NN+15\%MLM) &86.00 &89.33 &80.00 &80.00 &76.33 &69.67 &85.34 &85.24 &78.86 &90.00 &92.67 &88.00 &85.33 &85.89 &79.13 \\
100\% NE+ 15\% TLM on (100\%NE+100\%NN+15\%MLM) &86.00 &86.67 &80.00 &78.33 &76.67 &65.00 &82.55 &85.57 &78.52 &90.67 &93.00 &88.33 &84.39 &85.48 &77.96 \\
100\% VB+15\% TLM on (100\%NE+100\%NN+15\%MLM) &84.67 &87.67 &79.33 &80.67 &78.67 &70.00 &81.54 &85.23 &78.86 &89.33 &93.00 &87.00 &84.05 &86.14 &78.80 \\
100\% NN+ 15\% TLM on (100\%NE+100\%NN+15\%MLM) &85.67 &85.00 &78.00 &80.00 &76.33 &68.67 &81.21 &84.56 &77.18 &92.33 &93.00 &89.33 &84.80 &84.72 &78.29 \\
100\% NE+100\% VB+ 15\% TLM on (100\%NE+100\%NN+15\%MLM) &84.33 &85.33 &77.33 &79.00 &77.67 &69.00 &84.56 &85.91 &80.20 &91.00 &92.67 &89.00 &84.72 &85.39 &78.88 \\
100\% NE+100\% NN+ 15\% TLM on (100\%NE+100\%NN+15\%MLM) &86.67 &83.67 &78.67 &76.33 &78.00 &66.00 &82.55 &85.57 &78.19 &91.33 &91.67 &87.67 &84.22 &84.73 &77.63 \\
100\%NE+100\%VB+100\%NN+15\%TLM on (100\%NE+100\%NN+15\%MLM) &82.33 &86.00 &76.67 &77.00 &79.00 &68.67 &81.21 &82.55 &75.84 &91.00 &91.33 &88.00 &82.89 &84.72 &77.29 \\
\multicolumn{16}{c}{\textbf{}} \\
15\% TLM on (100\%NE+100\%VB+100\%NN+15\%MLM) &84.00 &86.00 &79.00 &83.00 &77.33 &71.67 &82.55 &85.23 &77.85 &90.33 &\colorbox{yellow}{\textbf{94.33}} &88.67 &84.97 &85.73 &79.30 \\
100\% NE+ 15\% TLM on (100\%NE+100\%VB+100\%NN+15\%MLM) &82.33 &86.33 &77.67 &80.00 &77.33 &68.67 &81.88 &84.56 &76.85 &88.67 &91.67 &86.67 &83.22 &84.97 &77.46 \\
100\% VB+15\% TLM on (100\%NE+100\%VB+100\%NN+15\%MLM) &84.67 &88.00 &79.67 &79.00 &77.00 &68.00 &83.89 &84.90 &78.52 &91.00 &94.00 &90.00 &84.64 &85.97 &79.05 \\
100\% NN+ 15\% TLM on (100\%NE+100\%VB+100\%NN+15\%MLM) &85.33 &87.00 &81.33 &77.00 &74.67 &66.33 &82.55 &84.90 &78.86 &89.33 &93.00 &87.33 &83.55 &84.89 &78.46 \\
100\% NE+100\% VB+ 15\% TLM on (100\%NE+100\%VB+100\%NN+15\%MLM) &82.67 &85.67 &78.33 &76.67 &75.00 &66.00 &83.22 &85.91 &77.52 &88.00 &92.67 &85.67 &82.64 &84.81 &76.88 \\
100\% NE+ 100\% NN+ 15\% TLM on (100\%NE+100\%VB+100\%NN+15\%MLM) &84.33 &84.00 &78.00 &76.67 &77.00 &67.33 &83.21 &85.57 &78.19 &87.00 &93.00 &85.00 &82.80 &84.89 &77.13 \\
100\%NE+100\%VB+100\%NN+15\%TLM on (100\%NE+100\%VB+100\%NN+15\%MLM) &85.33 &85.33 &79.67 &76.33 &76.00 &67.00 &82.22 &84.90 &77.12 &89.00 &94.00 &88.33 &83.22 &85.06 &78.03 \\
\bottomrule
\end{tabular}
}
\end{sidewaystable*}

\begin{sidewaystable*}[p]\centering
\caption{Ablation experiments and bitext mining scores for Sinhala-Tamil language pair considering linguistic entity masking.}\label{tab:sita_ablation}
\scriptsize
\resizebox{\linewidth}{!}{%
\begin{tabular}{lrrrrrrrrrrrrrrrr}\toprule
\multirow{2}{*}{\textbf{Experiment}} &\multicolumn{3}{c}{\textbf{Army}} &\multicolumn{3}{c}{\textbf{Hiru}} &\multicolumn{3}{c}{\textbf{ITN}} &\multicolumn{3}{c}{\textbf{Newsfirst}} &\multicolumn{3}{c}{\textbf{Average}} \\\cmidrule{2-16}
&\textbf{FW} &\textbf{BW} &\textbf{IN} &\textbf{FW} &\textbf{BW} &\textbf{IN} &\textbf{FW} &\textbf{BW} &\textbf{IN} &\textbf{FW} &\textbf{BW} &\textbf{IN} &\textbf{FW} &\textbf{BW} &\textbf{IN} \\\midrule
\textbf{Baselines} &\textbf{} &\textbf{} &\textbf{} &\textbf{} &\textbf{} &\textbf{} &\textbf{} &\textbf{} &\textbf{} &\textbf{} &\textbf{} &\textbf{} &\textbf{} &\textbf{} &\textbf{} \\
XLM-R &83.44 &81.46 &78.15 &90.67 &91.00 &87.33 &91.33 &90.00 &87.00 &\colorbox{yellow}{\textbf{93.67}} &95.33 &\colorbox{yellow}{\textbf{92.33}} &89.78 &89.45 &86.20 \\
15\%MLM &86.75 &88.08 &81.46 &88.00 &89.33 &84.00 &93.33 &92.67 &89.33 &90.33 &94.00 &89.00 &89.60 &91.02 &85.95 \\
15\%TLM on 15\%MLM &87.75 &90.40 &83.11 &88.67 &93.33 &86.33 &93.00 &94.33 &90.00 &91.33 &94.33 &89.67 &90.19 &93.10 &87.28 \\
\hline
\multicolumn{16}{l}{\textbf{LEM$_{mono}$}} \\
100\% NE+15\% MLM &86.42 &92.05 &83.78 &89.33 &92.00 &87.67 &94.00 &94.33 &90.67 &91.33 &94.00 &88.67 &90.27 &93.10 &87.69 \\
100\% VB+15\% MLM &83.44 &88.08 &78.81 &87.33 &90.33 &83.33 &92.33 &94.00 &88.00 &90.00 &92.00 &87.67 &88.28 &91.10 &84.45 \\
100\% NN+15\% MLM &85.10 &87.75 &80.13 &88.00 &91.67 &85.33 &92.00 &91.67 &88.00 &90.67 &93.33 &87.67 &88.94 &91.10 &85.28 \\
100\% NE+100\%VB+15\% MLM &84.43 &90.73 &82.12 &88.67 &91.00 &85.33 &94.00 &92.33 &88.33 &91.00 &94.33 &88.00 &89.53 &92.10 &85.95 \\
100\% NE+100\%NN+15\% MLM &85.43 &88.08 &79.47 &88.33 &89.67 &85.00 &95.00 &94.67 &91.33 &92.33 &93.33 &89.67 &90.27 &91.44 &86.37 \\
100\% NE+100\% VB +100\%NN+15\% MLM &83.11 &88.41 &79.80 &86.67 &91.33 &84.33 &91.67 &89.67 &85.33 &90.33 &93.67 &88.33 &87.94 &90.77 &84.45 \\
\hline
\multicolumn{16}{l}{\textbf{LEM$_{mono}$+LEM$_{para}$}}\\
100\% NE+15\% TLM on 15\%MLM &89.07 &90.73 &85.10 &89.33 &91.00 &85.67 &95.67 &94.67 &92.67 &91.00 &93.33 &88.33 &91.27 &92.43 &87.94 \\
100\% VB+15\% TLM on 15\%MLM &88.41 &91.00 &84.77 &87.67 &91.00 &85.67 &93.67 &93.67 &90.67 &92.33 &93.00 &90.00 &90.52 &92.17 &87.78 \\
100\% NN+15\% TLM on 15\%MLM &88.74 &90.07 &84.44 &89.67 &91.67 &86.67 &94.67 &93.33 &90.67 &92.00 &91.67 &87.33 &91.27 &91.68 &87.28 \\
100\% NE+100\% VB+15\% TLM on 15\%MLM &86.75 &90.73 &83.11 &89.67 &90.33 &86.33 &92.67 &92.67 &88.67 &92.33 &\colorbox{yellow}{\textbf{95.33}} &90.67 &90.36 &92.27 &87.19 \\
100\% NE+100\% NN+15\% TLM on 15\%MLM &86.42 &90.73 &83.11 &87.33 &89.33 &84.00 &94.67 &93.33 &91.67 &92.00 &93.67 &88.33 &90.11 &91.77 &86.78 \\
100\% NE+100\% VB+ 100\% NN+ 15\% TLM on 15\%MLM &85.43 &91.39 &81.79 &89.00 &92.33 &86.67 &93.67 &93.33 &88.67 &90.33 &93.00 &87.00 &89.61 &92.51 &86.03 \\
\multicolumn{16}{c}{\textbf{}} \\
15\% TLM on 100\%NE+15\%MLM &87.09 &89.73 &83.44 &89.33 &92.00 &86.33 &94.33 &92.67 &89.33 &92.00 &93.67 &89.67 &90.69 &92.02 &87.19 \\
15\% NE+15\%TLM on 100\%NE+15\%MLM &88.33 &\colorbox{yellow}{\textbf{93.33}} &87.33 &88.33 &93.33 &87.33 &93.33 &94.00 &89.00 &92.00 &93.67 &89.67 &90.50 &\colorbox{yellow}{\textbf{93.58}} &88.33 \\
100\% VB+15\% TLM on 100\%NE+15\%MLM &86.42 &90.07 &83.11 &90.00 &92.00 &87.67 &94.33 &93.00 &90.33 &92.00 &94.67 &90.00 &90.69 &92.43 &87.78 \\
100\% NN+15\% TLM on 100\%NE+15\%MLM &86.09 &91.72 &83.78 &89.67 &92.67 &87.67 &95.33 &95.00 &91.67 &92.67 &93.33 &89.67 &90.94 &93.18 &88.19 \\
100\% NE+100\% VB+15\% TLM on 100\%NE+15\%MLM/ &87.09 &90.07 &84.11 &89.00 &91.33 &86.67 &95.67 &94.67 &91.67 &90.67 &92.33 &88.67 &90.61 &92.10 &87.78 \\
100\% NE+100\% NN+15\% TLM on 100\%NE+15\%MLM/ &86.09 &90.73 &83.11 &90.00 &\colorbox{yellow}{\textbf{93.67}} &\colorbox{yellow}{\textbf{89.33}} &94.33 &94.00 &90.33 &91.00 &\colorbox{yellow}{\textbf{95.33}} &89.33 &90.36 &93.43 &88.03 \\
100\%NE+100\%VB+100\%NN+15\%TLM on 100\%NE+15\%MLM/ &86.09 &93.05 &84.11 &89.67 &92.00 &88.00 &95.67 &95.00 &\colorbox{yellow}{\textbf{93.33}} &91.00 &94.00 &89.33 &90.61 &93.51 &88.69 \\
\multicolumn{16}{c}{\textbf{}} \\
15\% TLM on (100\%VB+15\%MLM) &89.07 &88.41 &83.44 &89.67 &92.33 &87.00 &93.33 &93.67 &90.00 &91.00 &91.67 &87.33 &90.77 &91.52 &86.94 \\
100\% NE+ 15\% TLM on (100\%VB+15\%MLM) &87.75 &88.74 &83.11 &90.00 &91.00 &86.33 &95.00 &94.67 &91.67 &93.00 &91.67 &88.00 &91.44 &91.52 &87.28 \\
100\% VB+ 15\% TLM on (100\%VB+15\%MLM) &88.74 &90.75 &84.44 &89.67 &92.00 &86.67 &92.67 &94.33 &89.33 &92.33 &93.33 &89.33 &90.85 &92.60 &87.44 \\
100\% NN+15\% TLM on (100\%VB+15\%MLM) &86.42 &90.73 &84.11 &89.67 &92.33 &86.33 &91.33 &93.33 &88.00 &92.00 &94.00 &89.67 &89.86 &92.60 &87.03 \\
100\% NE+ 100\% VB+ 15\% TLM on (100\%VB+15\%MLM) &85.76 &88.08 &81.13 &90.33 &92.33 &88.00 &93.00 &91.33 &88.67 &92.33 &92.00 &88.33 &90.36 &90.94 &86.53 \\
100\% NE+ 100\% NN+ 15\% TLM on (100\%VB+15\%MLM) &87.09 &89.07 &82.78 &88.67 &92.33 &85.00 &93.33 &93.67 &89.67 &92.00 &91.33 &87.00 &90.27 &91.60 &86.11 \\
100\% NE+ 100\% NN+ 100\%VB+ 15\% TLM on (100\%VB+15\%MLM) &85.77 &90.40 &83.11 &89.67 &92.00 &86.00 &92.33 &93.67 &88.67 &92.00 &92.00 &88.00 &89.94 &92.02 &86.44 \\
\multicolumn{16}{c}{\textbf{}} \\
15\% TLM on (100\%NN+15\%MLM) &88.41 &91.39 &85.76 &88.33 &92.67 &85.67 &95.67 &\colorbox{yellow}{\textbf{95.67}} &91.67 &91.00 &93.67 &89.33 &90.85 &93.35 &88.11 \\
100\% NE+ 15\% TLM on (100\%NN+15\%MLM) &\colorbox{yellow}{\textbf{89.40}} &92.72 &87.42 &90.13 &93.33 &88.67 &96.67 &93.00 &90.67 &92.33 &93.67 &89.67 &\colorbox{yellow}{\textbf{92.13}} &93.18 &89.10 \\
100\% VB+15\% TLM on (100\%NN+15\%MLM) &87.75 &90.07 &83.11 &87.67 &92.00 &85.00 &93.33 &93.33 &89.00 &89.67 &92.33 &87.67 &89.60 &91.93 &86.19 \\
100\% NN+ 15\% TLM on (100\%NN+15\%MLM) &85.43 &90.73 &82.12 &88.67 &93.00 &86.67 &95.33 &93.67 &91.00 &93.00 &92.67 &89.33 &90.61 &92.52 &87.28 \\
100\% NE+ 100\% VB+ 15\% TLM on (100\%NN+15\%MLM) &86.09 &90.40 &82.78 &\colorbox{yellow}{\textbf{91.33}} &92.00 &88.33 &94.67 &94.67 &91.33 &91.33 &93.00 &88.33 &90.86 &92.52 &87.69 \\
100\% NE+ 100\% NN+ 15\% TLM on (100\%NN+15\%MLM) &87.75 &89.40 &83.11 &88.33 &92.00 &85.67 &95.00 &93.67 &90.67 &92.00 &93.00 &89.33 &90.77 &92.02 &87.19 \\
100\% NE+ 100\% NN+ 100\%VB+ 15\% TLM on (100\%NN+15\%MLM) &85.43 &92.05 &83.78 &89.33 &93.00 &87.00 &94.33 &93.33 &90.00 &90.67 &92.67 &88.00 &89.94 &92.76 &87.19 \\
\multicolumn{16}{c}{\textbf{}} \\
15\% TLM on (100\%NE+100\%VB+15\%MLM) &86.09 &91.06 &83.44 &90.33 &90.67 &87.33 &96.33 &94.33 &92.33 &92.33 &94.33 &90.00 &91.27 &92.60 &88.28 \\
100\% NE+ 15\% TLM on (100\%NE+100\%VB+15\%MLM) &86.75 &90.07 &83.44 &89.33 &91.33 &86.67 &93.67 &94.67 &91.33 &92.00 &93.33 &89.67 &90.44 &92.35 &87.78 \\
100\% VB+15\% TLM on (100\%NE+100\%VB+15\%MLM) &84.44 &91.39 &81.46 &89.33 &91.67 &85.00 &95.33 &93.33 &91.00 &92.00 &93.33 &89.33 &90.28 &92.43 &86.70 \\
100\% NN+ 15\% TLM on (100\%NE+100\%VB+15\%MLM) &85.76 &92.05 &85.00 &90.67 &93.00 &88.33 &95.67 &95.33 &92.67 &89.33 &92.67 &86.33 &90.36 &93.26 &88.08 \\
100\% NE+100\% VB+ 15\% TLM on (100\%NE+100\%VB+15\%MLM) &87.09 &89.73 &83.11 &90.67 &92.33 &88.33 &94.67 &92.67 &89.33 &92.00 &93.00 &90.00 &91.10 &91.93 &87.69 \\
100\% NE+ 100\% NN+ 15\% TLM on (100\%NE+100\%VB+15\%MLM) &85.76 &92.05 &85.00 &90.67 &91.67 &87.67 &95.67 &94.00 &91.33 &90.67 &93.67 &88.33 &90.69 &92.85 &88.08 \\
100\%NE+100\%VB+100\%NN+15\%TLM on (100\%NE+100\%VB+15\%MLM) &86.75 &92.05 &84.77 &88.33 &90.00 &86.67 &94.33 &94.67 &90.67 &89.67 &93.67 &87.67 &89.77 &92.60 &87.44 \\
\multicolumn{16}{c}{\textbf{}} \\
15\% TLM on (100\%NE+100\%NN+15\%MLM) &86.75 &91.72 &83.11 &90.67 &92.33 &88.67 &95.67 &93.67 &91.33 &92.00 &94.33 &90.00 &91.27 &93.01 &88.28 \\
100\% NE+ 15\% TLM on (100\%NE+100\%NN+15\%MLM) &86.75 &87.47 &81.79 &90.33 &93.00 &88.00 &95.33 &93.33 &90.33 &\colorbox{yellow}{\textbf{93.67}} &94.00 &90.33 &91.52 &91.95 &87.61 \\
100\% VB+15\% TLM on (100\%NE+100\%NN+15\%MLM) &87.75 &90.40 &83.44 &89.33 &92.33 &86.67 &95.00 &94.00 &91.33 &91.67 &94.67 &89.67 &90.94 &92.85 &87.78 \\
100\% NN+ 15\% TLM on (100\%NE+100\%NN+15\%MLM) &87.75 &90.40 &84.77 &87.67 &89.67 &84.00 &95.33 &95.33 &92.00 &90.00 &93.67 &88.67 &90.19 &92.27 &87.36 \\
100\% NE+100\% VB+ 15\% TLM on (100\%NE+100\%NN+15\%MLM) &85.76 &87.75 &81.13 &90.33 &90.67 &86.00 &94.33 &92.00 &89.00 &92.33 &93.67 &90.00 &90.69 &91.02 &86.53 \\
100\% NE+100\% NN+ 15\% TLM on (100\%NE+100\%NN+15\%MLM) &87.75 &90.07 &84.44 &90.33 &92.00 &86.67 &96.00 &94.00 &91.67 &93.00 &93.67 &89.00 &91.77 &92.43 &87.94 \\
100\%NE+100\%VB+100\%NN+15\%TLM on (100\%NE+100\%NN+15\%MLM) &85.76 &88.74 &\colorbox{yellow}{\textbf{91.46}} &89.67 &92.00 &86.00 &94.33 &93.67 &91.00 &93.00 &94.00 &89.67 &90.69 &92.10 &\colorbox{yellow}{\textbf{89.53}} \\
\multicolumn{16}{c}{\textbf{}} \\
15\% TLM on (100\%NE+100\%VB+100\%NN+15\%MLM) &86.09 &89.40 &82.45 &89.33 &92.00 &87.00 &94.33 &91.67 &88.33 &91.33 &92.00 &86.33 &90.27 &91.27 &86.03 \\
100\% NE+ 15\% TLM on (100\%NE+100\%VB+100\%NN+15\%MLM) &88.76 &89.40 &84.44 &89.67 &91.33 &87.33 &95.00 &94.00 &90.67 &90.33 &92.00 &87.00 &90.94 &91.68 &87.36 \\
100\% VB+15\% TLM on (100\%NE+100\%VB+100\%NN+15\%MLM) &85.76 &89.40 &82.12 &90.33 &93.33 &88.67 &94.67 &94.00 &90.67 &93.00 &93.67 &90.00 &90.94 &92.60 &87.86 \\
100\% NN+ 15\% TLM on (100\%NE+100\%VB+100\%NN+15\%MLM) &85.43 &90.73 &82.78 &89.67 &91.67 &87.33 &95.33 &92.33 &90.33 &90.00 &93.33 &86.67 &90.11 &92.02 &86.78 \\
100\% NE+100\% VB+ 15\% TLM on (100\%NE+100\%VB+100\%NN+15\%MLM) &86.42 &91.00 &82.78 &88.67 &91.67 &86.33 &94.00 &92.67 &89.33 &93.00 &93.67 &90.33 &90.52 &92.25 &87.19 \\
100\% NE+ 100\% NN+ 15\% TLM on (100\%NE+100\%VB+100\%NN+15\%MLM) &86.75 &90.73 &84.11 &90.67 &93.33 &88.33 &\colorbox{yellow}{\textbf{97.33}} &92.67 &92.33 &91.33 &92.00 &86.67 &91.52 &92.18 &87.86 \\
100\%NE+100\%VB+100\%NN+15\%TLM on (100\%NE+100\%VB+100\%NN+15\%MLM) &87.42 &89.40 &82.78 &89.33 &92.33 &86.33 &94.67 &95.00 &91.33 &91.33 &94.00 &88.67 &90.69 &92.68 &87.28 \\
\bottomrule
\end{tabular}
}
\end{sidewaystable*}

\clearpage

\section{Limitations of the NER Model and Pos Tagger}\label{secD}

Examples highlighting the error categories found with the NER model and PoS taggers (Section~\ref{sec:discussion}) are shown in Table~\ref{tab:ner_limitations} and Table~\ref{tab:pos_limitations} respectively. 

\begin{table}[!htb]
    \centering
    \caption{Examples of incorrect identification and labeling of NEs. We identify mainly two error categories: false positives and false negatives, where the NER model underperforms.}\label{tab:ner_limitations}
    \begin{tabular}{c}
    \includegraphics[width=\linewidth]{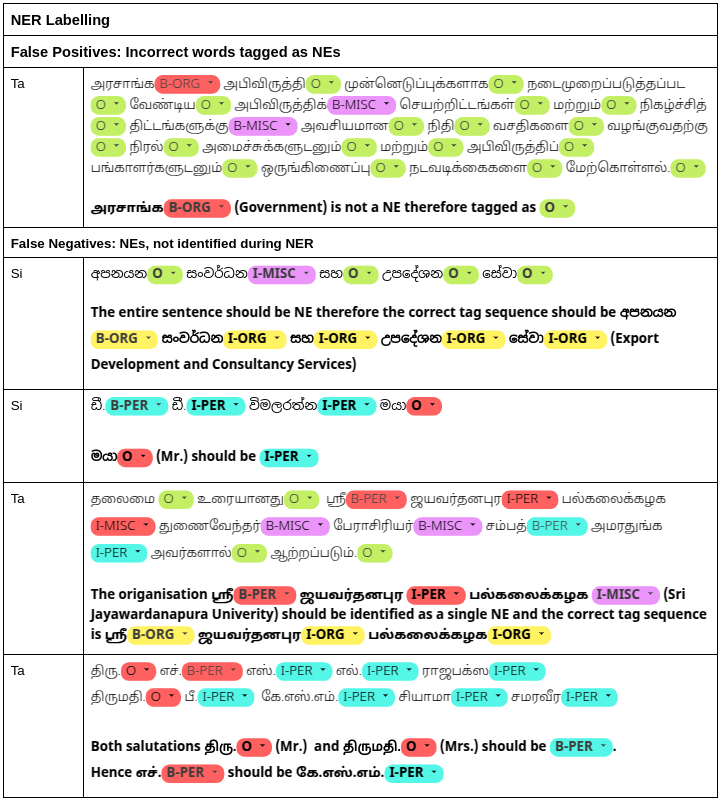} \\
    \end{tabular}
\end{table}

\begin{table}[!htb]
    \centering
    \caption{Examples of incorrect identification and labelling of PoS Tags. We identify mainly two error categories: false positives and false negatives, where the Pos Tagger underperforms.}\label{tab:pos_limitations}
    \begin{tabular}{c}
    \includegraphics[width=\linewidth]{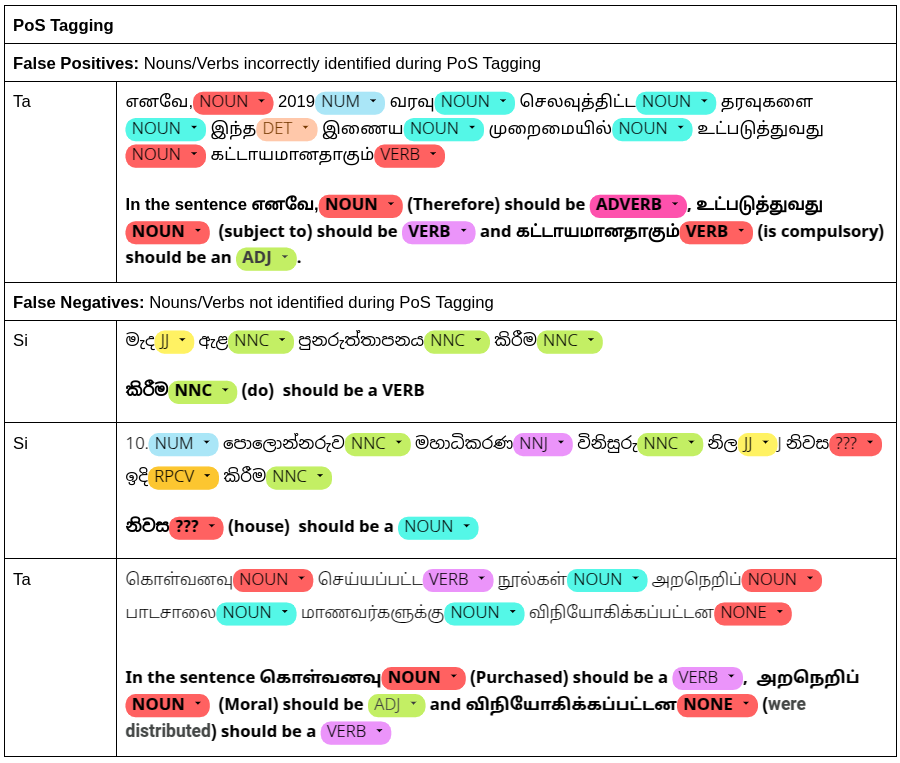} \\
    \end{tabular}
\end{table}

\end{appendices}
\clearpage
\bibliography{sn-bibliography}

\end{document}